\documentclass{article}

\usepackage[preprint]{neurips_2026}
\usepackage{microtype}
\usepackage{graphicx}
\usepackage{subcaption}
\usepackage{booktabs} %
\newcommand{\quotes}[1]{``#1''}
\usepackage{wrapfig}

\usepackage[utf8]{inputenc} %
\usepackage[T1]{fontenc}    %
\usepackage{hyperref}       %
\usepackage{url}            %
\usepackage{booktabs}       %
\usepackage{amsfonts}       %
\usepackage{nicefrac}       %
\usepackage{microtype}      %
\usepackage{xcolor}         %
\usepackage{amsmath}
\usepackage[dvipsnames]{xcolor}

\title{Quantum Implicit Neural Representations for\\Novel View Synthesis}

\author{
Daniele {Lizzio Bosco}$^{1,2}$\thanks{work done during an internship at the 4DQV research group, MPI for Informatics} \quad
Shuteng Wang$^{3}$ \quad
Giuseppe Serra$^{1}$ \quad
Vladislav Golyanik$^{3,}$\thanks{corresponding author’s email: \texttt{golyanik@mpi-inf.mpg.de}}
 \\
\\
$^{1}$Department of Mathematics, Computer Science and Physics, University of Udine, Italy \\
$^{2}$Department of Biology, University of Naples Federico II, Italy \\
$^{3}$Max Planck Institute for Informatics, SIC, Saarbrücken, Germany \\
\\\vspace{10pt}
\definecolor{hblue}{HTML}{2F6DA1}
\textcolor{hblue}{\url{https://4dqv.mpi-inf.mpg.de/3D-QISR/}}
}

\begin{document}

\maketitle

\begin{abstract}
Quantum implicit neural representations have recently emerged as compact function approximators for continuous visual signals.
In parallel, Neural Radiance Fields~(NeRFs) have become a dominant framework for novel-view synthesis by recovering continuous volumetric scene representations from posed 2D images, but often require substantial model capacity and intensive optimisation.
We introduce \textbf{3D Quantum Implicit Scene Representation} (3D-QISR), the first hybrid quantum-classical radiance-field model for novel-view synthesis from 2D observations.
3D-QISR replaces the classical NeRF backbone with parameterised quantum circuits that encode spatial and view-dependent information through quantum state transformations, enabling compact quantum-enhanced implicit representations while preserving compatibility with standard volumetric rendering.
We propose two architectures: \textbf{Full 3D-QISR}, which maximises the representational role of a unified quantum state, and \textbf{Dual-Branch 3D-QISR}, which separates spatial and view-dependent quantum embeddings to introduce a task-aligned inductive bias, reduce state-preparation complexity, and improve scalability and hardware compatibility.
Experiments on moderate-resolution novel-view synthesis benchmarks show that 3D-QISR executed on simulated quantum hardware matches or outperforms classical baselines while using fewer than half the trainable parameters.
These results position quantum neural networks as compact and competitive building blocks for continuous volumetric scene representation, opening a new direction for quantum-enhanced implicit neural rendering. 
\end{abstract} 

\section{Introduction} 
Implicit neural representations~(INRs) have become a central paradigm for modelling continuous visual signals with coordinate-based learning models. 
Among them, Neural Radiance Fields~(NeRFs) have had a particularly strong impact on novel-view synthesis by representing a 3D scene as a continuous volumetric function, typically parameterised by a multi-layer perceptron~(MLP), that maps spatial locations and viewing directions to density and radiance ~\citep{NERF, mip-NERF, survey_nerf}. 
NeRF-based models have since been extended to a broad range of visual computing tasks, including 3D reconstruction \citep{NERF-, NARF, MVSNERF}, dynamic and human-centric modelling \citep{nerfies, implicitNeRF, graspNERF}, and image processing \citep{layeredNeRF, imageprocessingNERF}. 
Despite this progress, their performance remains strongly tied to the representational and optimisation properties of the underlying neural field; improving reconstruction quality or convergence speed often requires increasing network capacity, enriching positional encodings, or introducing task-specific architectural modifications. 
This paper investigates whether Quantum Machine Learning (QML) can provide an alternative foundation for compact and trainable implicit representations of continuous 3D visual fields. 
Quantum neural networks~(QNNs) define a distinct model class in which data are embedded into quantum states, transformed by trainable unitary operations, and read out through quantum measurements \citep{PowerofQNNs}. 
Through superposition, interference, and entanglement, QNNs naturally induce high-dimensional feature transformations, offering a potentially compact mechanism for representing complex functions. 
Motivated by theoretical and empirical studies on the expressivity, learnability, and compactness of quantum models \citep{PowerofQNNs, quantum_kernel_advantage, advantage_separability}, recent work has begun to explore quantum models as function approximators for visual signals~\citep{Quantumimplicitneuralrepresentations, Wang2025QVFs}.
However, extending quantum implicit representations from fully supervised field regression to NeRF-style novel-view synthesis is non-trivial: NeRF training is an inverse rendering problem, where the 3D density and radiance fields must be recovered indirectly from 2D image observations through differentiable volumetric rendering.
This setting introduces additional challenges for quantum models, including coordinate-to-state embedding, measurement design, output scaling, noise sensitivity, and trainability under sparse and view-dependent supervision.
As a result, it remains unclear whether quantum neural networks can act as effective building blocks for volumetric 3D reconstruction from 2D images. 
We address this gap by introducing a \textbf{3D Quantum Implicit Scene Representation}~(\textbf{3D-QISR})
for novel-view synthesis from 2D observations.
Inspired by the volumetric scene formulation and differentiable rendering pipeline of NeRF, 3D-QISR departs from existing Quantum Implicit Neural Representations \citep{Quantumimplicitneuralrepresentations, Wang2025QVFs} (QINRs) by learning a compact 3D representation indirectly from posed images rather than fitting a fully observed field.
Specifically, spatial coordinates and viewing directions are mapped by a learnable classical embedding into quantum states, transformed by parameterised quantum circuits, and measured to produce radiance-field quantities such as density and colour, which are then integrated through standard volume rendering.
This design places QML-based containers at the core of the scene representation while retaining the image-based supervision mechanism required for inverse 3D reconstruction.
We instantiate the framework in two architectures: \textbf{Full 3D-QISR}, which uses a unified quantum representation to maximise the role of the quantum circuit, and \textbf{Dual-Branch 3D-QISR}, which separates spatial and view-dependent inputs into distinct quantum branches, introducing a task-aligned inductive bias for decomposing geometry and appearance while improving scalability and robustness under simulated hardware noise.
Our contributions are summarised as follows:
\begin{itemize}
    \item \textbf{3D-QISR}, the first quantum-enhanced implicit scene representation for novel-view synthesis from posed 2D images for gate-based quantum computers (Sec.~\ref{sec:Method}); 
    \item A \textbf{Dual-Branch 3D-QISR architecture} that separately encodes spatial and view-dependent information, yielding a task-aligned inductive bias for quantum implicit 3D representation learning (Sec.~\ref{sub:MPLbased});
    \item Design components for trainable 3D-QISRs (Sec.~\ref{sub:quantum_design}), including 
    local measurement schemes (Sec.~\ref{ssec:parity_based_measurement}) and output scaling (Sec.~\ref{ssec:output_scaling}), which improve optimisation stability and robustness. 
\end{itemize} 
Experiments on two standard novel-view synthesis datasets~\citep{NERF,llff}, downscaled to enable tractable quantum simulation, show that \textbf{Full 3D-QISR} achieves higher reconstruction quality than classical NeRF-based baselines while using fewer than half the trainable parameters. 
\textbf{Dual-Branch 3D-QISR} attains comparable reconstruction quality and exhibits stronger robustness under simulated quantum noise.
We further evaluate 3D-QISR under multiple inference-time noise models and provide additional analyses in the Appendix, including noise-resilience studies, meshes extracted from trained 3D-QISRs, and compatibility with mip-NeRF-style extensions~\citep{mip-NERF}.
Overall, these results position QNNs as compact and competitive building blocks for continuous volumetric scene representation, opening a new direction for quantum-enhanced implicit neural rendering.
To encourage further research in this field, we will release the full implementation of our model along with the experimental codes and training details.

\section{Related Work}

\paragraph{Classical 3D Scene Representation for Novel-view Synthesis.} 
Neural networks have long been applied to image-based view synthesis \citep{DeepView, liu2019softras, llff}.
However, a breakthrough in the field was the introduction of Neural Radiance Fields (NeRF) \citep{NERF}, which proposed learning a continuous volumetric scene representation from posed images. NeRF defines a 5D mapping from spatial location and viewing direction to colour and density, enabling high-fidelity novel view synthesis. The success of NeRF set a new standard for rendering quality and inspired a substantial body of subsequent research \citep{survey_NeuralFieldsVisualComputing, survey_nerf, yariv2023bakedsdf, li2023dynibar, mildenhall2022rawnerf}.
Numerous extensions have targeted improvements in rendering performance, generalisation, and robustness under specific conditions, such as mip-NeRF \citep{mip-NERF}, which improved performance at multiple resolutions, and instant-NGP \citep{instant-ngp},  developed to reduce training time. It is worth noting that many of these NeRF variants retain the core architectural element of an MLP to learn the radiance field, and could, therefore, benefit from a hybrid classical-quantum architecture such as the one we propose in this work. Since our proposed quantum architecture enhances the \quotes{core building block} of NeRF, these variants are only partially related to this work. We rethink the foundations of 3D scene representation learning from 2D observations. 
\paragraph{Quantum Implicit Representations.} 
Quantum-enhanced Computer Vision (QeCV) has recently emerged at the intersection of quantum machine learning and computer vision, focusing on leveraging quantum computing for tasks in image processing, graphics, and 3D reconstruction \citep{KueteMeli2025arXiv}. Recent work includes quantum convolutional neural networks \citep{Quanv, qccnn, qccnn2}, quantum generative adversarial networks \citep{qgan1, qgan2}, quantum autoencoders \citep{Rathi2023}, and other hybrid architectures  \citep{image_classification_AE, quantum_vision_transformer, QNN_for_classification}.
More recently, Quantum Implicit Neural Representations~(QINRs) have been proposed to extend coordinate-based implicit learning by replacing or augmenting classical neural-field containers with quantum models. Typically, coordinates are embedded into quantum states, transformed by parameterised quantum circuits, and decoded through quantum measurements, yielding continuous function approximators whose features arise from quantum state evolution. In particular, QIREN \citep{Quantumimplicitneuralrepresentations} demonstrated the use of QNNs for the task of super-resolution (from greyscale images of size $32\times 32$), while Quantum Visual Fields \citep{Wang2025QVFs} (QVF) obtained promising results in the reconstruction of 2D images (evaluated on collections of CIFAR $32\times 32$px images, and on a single image of $400\times 400$px).

These works have shown that QINRs have the ability to compactly fit visual signals and capture high-frequency structure with favourable parameter efficiency~\citep{Quantumimplicitneuralrepresentations,Wang2025QVFs}. These properties are highly relevant to neural rendering, where the field continues to seek scene representations that are expressive, trainable, and compact. The ability of QINRs to overfit continuous visual signals is advantageous for scene-specific reconstruction, where each model is optimised to represent a single scene. However, existing QINRs remain limited to the same-dimensional supervised field regression, such as fitting a 2D image from 2D coordinates or a 3D signal from observed 3D samples. They do not address the inverse problem central to novel-view synthesis, where a 3D representation must be inferred indirectly from posed 2D images through volumetric rendering. Our work extends QINRs to this setting by introducing a quantum implicit scene representation trained from image supervision. This requires task-specific innovations: (i) a radiance-field formulation that jointly processes spatial and view-dependent features; (ii) a dual-branch quantum encoding that separates position and viewing direction, reducing amplitude-embedding and state-preparation complexity~\citep{complexityQuantumStatePrep} while introducing a geometry--appearance inductive bias; and (iii) output scaling and local measurements to mitigate exponential concentration and trainability issues in quantum models~\citep{exponential_conc, BP, BP_review}.

\section{Quantum Machine Learning} 
This section provides the necessary background on Quantum Machine Learning (QML). We defer notation details to App.~\ref{app:bg_qc}. Comprehensive introductions to quantum computing and QML can be found in \cite{intro_to_QC}, \cite{intro_to_QML} and \cite{KueteMeli2025arXiv}. 
QML arises at the intersection of quantum computing and machine learning. The field encompasses a broad variety of models, methodologies, and objectives \citep{QML_review1, QML_review2}. One of the central goals of QML is to design models that exploit quantum phenomena such as entanglement and superposition to achieve an ``advantage'' in terms of computational resources, representational power, or learning performance when compared to purely classical methods \citep{ChallengesandOpportunities}.
A prominent class of models within QML is that of Quantum Neural Networks (QNNs). In \citep{PowerofQNNs}, it was proved that QNNs, in theory and in certain scenarios, can achieve higher effective dimensions than comparable classical neural networks, which translates into faster convergence during training. Following this theoretical result, experimental results from both general \citep{exp_eval_qml}  and vision-specific tasks \citep{quantum_vision_transformer, QNN_for_classification} show that QNNs can achieve the same-level or better performance than classical neural networks with fewer parameters and with increased convergence speed.
Given these properties, it is natural to investigate the application of QNNs to mid-level vision tasks such as 3D reconstruction and novel-view rendering, 
where efficient and faster training and generalisation are particularly critical due to the high dimensionality and complexity of the models required.
\paragraph{Quantum Neural Networks.}
In Quantum Machine Learning, the term Quantum Neural Network is often used interchangeably with Parameterised Quantum Circuit (PQC), as the two notions are closely related \citep{QNNs_as_PQC}. A QNN typically consists of a sequence of quantum gates whose operations depend on the classical inputs $\mathbf{x}$ and on a set of free parameters $\mathbf{\theta}$, which are optimised during the training process.
Given an initial state $|\phi\rangle$ (e.g. the state $|0\rangle^{\otimes N}$, where $N$ is the size of the quantum system), the application of a parameterised quantum circuit $P$ results in the state $P(\mathbf{\theta};\mathbf{x})|\phi\rangle$.
In many architectures, the circuit $P$ can be naturally decomposed into two stages: a data encoding stage and a variational (trainable) stage. The encoding is typically achieved through a fixed set of gates that map the classical input into a quantum state (often referred to as the \textit{embedding} or \textit{feature map}), denoted by $S(\mathbf{x})$. The variational stage, represented by $V(\mathbf{\theta})$, then acts on the embedded state.
Thus, the final state of the circuit can be described as $V(\mathbf{\theta})S(\mathbf{x})|\phi\rangle = V(\mathbf{\theta})|\phi_S(\mathbf{x})\rangle$, where $|\phi_S(\mathbf{x})\rangle$ denotes the quantum state obtained by embedding the input $\mathbf{x}$. The structure $V(\mathbf{\theta})$ typically constitutes a variational ansatz, designed to be expressive enough to represent the target function during learning.
\paragraph{Noisy Intermediate-Scale Quantum Era.}
Current quantum hardware operates in what is referred to as the \textit{Noisy Intermediate-Scale Quantum} (NISQ) era \citep{NISQ_preskill}. In this regime, quantum devices are composed of just tens to hundreds of qubits, and they are subject to non-negligible levels of noise and decoherence, which significantly limit their computational capabilities.
As a consequence, it is of fundamental importance to design quantum circuits that minimise both the number of quantum gates and the overall circuit depth. Excessive gate count or depth can result in an accumulation of errors that rapidly degrade the fidelity of the computation, making the outcomes unreliable. This limitation motivates the development of shallow, hardware-efficient quantum circuits and encourages a careful trade-off between expressivity and robustness to noise.

\section{3D Quantum Implicit Scene Representation (3D-QISR)} 
\label{sec:Method}

\begin{figure}[htbp]
    \centering
    \begin{subfigure}[t]{0.48\textwidth}
        \centering
        \includegraphics[width=\linewidth, trim=0 0 0.5cm 0, clip]{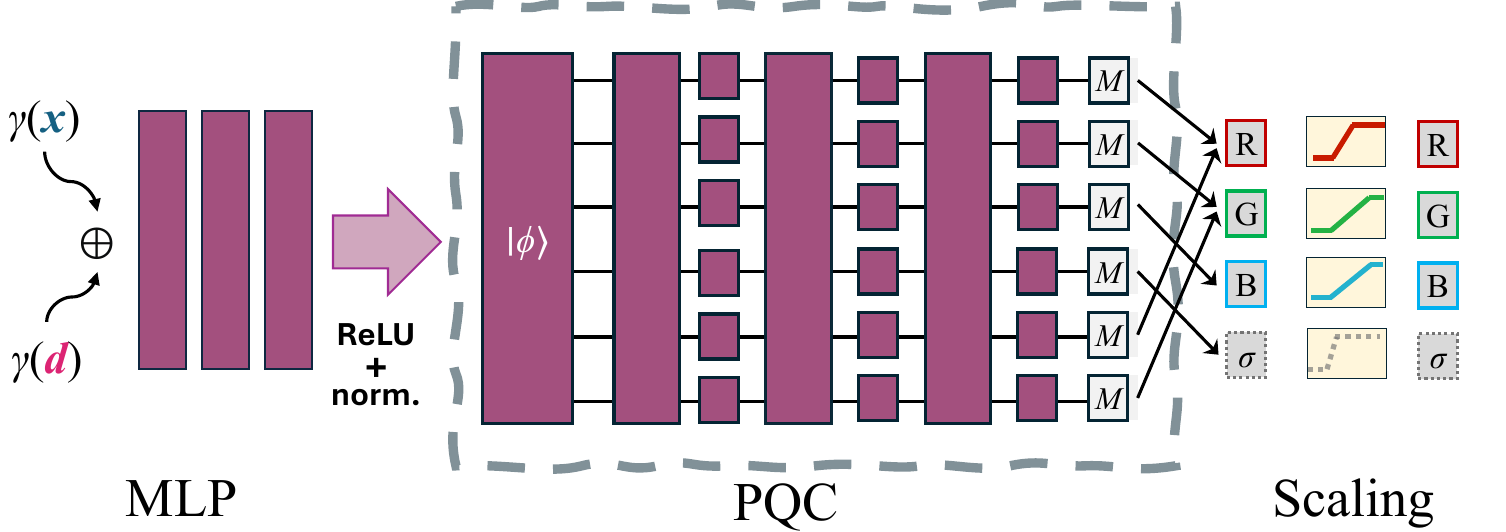}
        \caption{Full 3D-QISR}
        \label{fig:visual_method_a}
    \end{subfigure}
    \begin{subfigure}[t]{0.48\textwidth}
        \centering
        \includegraphics[width=\linewidth, trim=0 0 1cm 0, clip]{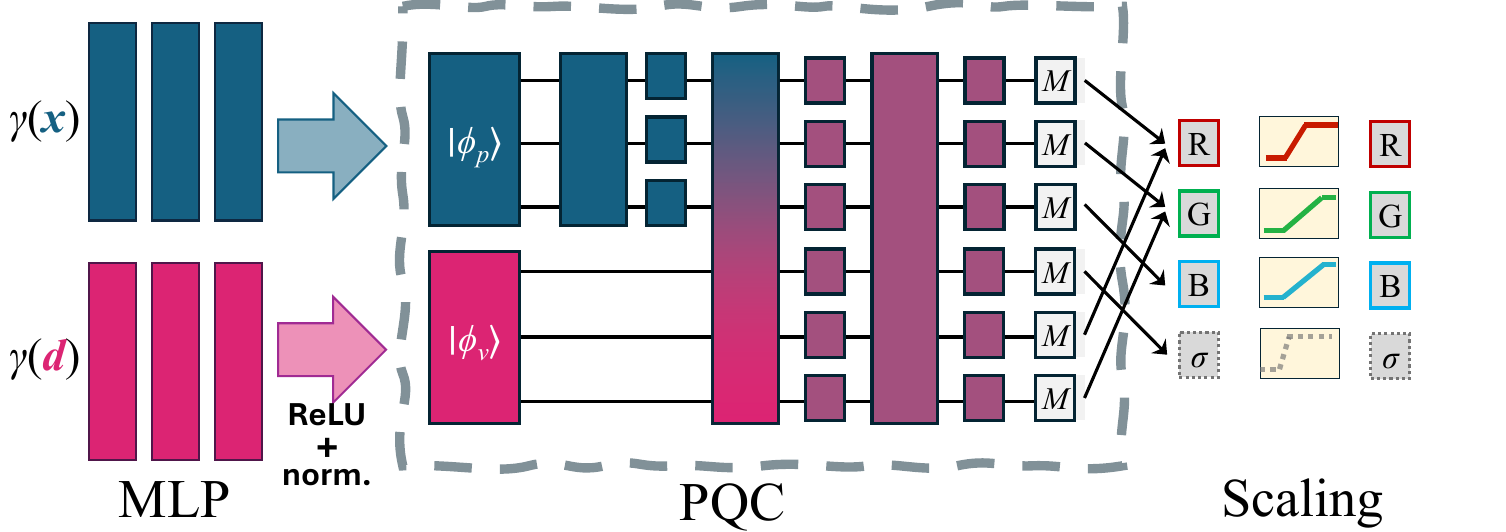}
        \caption{Dual-Branch 3D-QISR}
        \label{fig:visual_method_b}
    \end{subfigure}
    \caption{
    Scheme of the proposed models for $n=6$ qubits and $\ell=2$ repetitions. Positional and view-dependent coordinates (teal and magenta) are first encoded into quantum amplitudes by one (a) or two (b) MLPs, then processed by a PQC (dashed box). 
 Gates are coloured according to the information processed: purple (combined positional and view-dependent), blue or magenta (positional-only or view-dependent-only). 
    The quantum state is then measured using parity-based readout and finally scaled to reconstruct the output view.
    }
    \label{fig:visual_method}
\end{figure}

We describe our proposed method \textbf{3D Quantum Implicit Scene Representation} (3D-QISR), consisting of a hybrid quantum-classical model designed to efficiently represent scenes and generate novel views.  
A brief background on Neural Radiance Fields is provided in App.~\ref{app:bg_nerf}.
We start by describing how to encode classical coordinates into a quantum state. This description is given in two variants: one for the \textit{Full} encoding, which allows to leverage all the properties of the quantum system, and one for the \textit{Dual-Branch} encoding, which reduces the expressivity of the model to drastically increase scalability and compatibility with current quantum devices. A scheme of the proposed circuits is provided in Fig. \ref{fig:visual_method}. 
Then, we present the quantum circuit design choices. Finally, we describe postprocessing operations, consisting of a parity measurement to extract information, and then a \quotes{de-concentration} scaling layer that enhances the performance of the model by mitigating exponential concentration \citep{exponential_conc}.

\subsection{MLP-based Quantum Embedding}
\label{sub:MPLbased}
Neural Radiance Field \citep{NERF} models heavily rely on positional encodings for both spatial and view-dependent information. This technique enriches each input coordinate \( x \in \mathbb{R} \) by mapping it to a higher-dimensional space \( \mathbb{R}^{2L} \) using the transformation
\begin{equation}
    \gamma(x) = \left( \sin(2^0 \pi x), \dots, \cos(2^{L-1} \pi x) \right).
\end{equation}
This encoding has been shown to be crucial for mitigating the spectral bias of neural networks, which tend to under-represent high-frequency components, thereby enabling the representation of fine-grained variations in the data.
Even for modest values of $L$, the dimension of $\gamma(x)$ exceeds the number of qubits available in current quantum devices. Therefore, to obtain a quantum state that represents the structure from the encoded vector, we employ amplitude embedding.  Unlike angle embedding or unitary amplitude encoding \citep{unitary_amplitude_embedding}, which scale linearly with the number of features, amplitude embedding offers exponential compression, as a vector with $2^n$ components is represented using only $n$ qubits. 
\begin{wrapfigure}[8]{r}{0.5\columnwidth}
    \centering

    \begin{subfigure}[t]{0.24\textwidth}
        \centering
        \includegraphics[width=\linewidth, trim=0 3.75cm 15.5cm 0, clip]{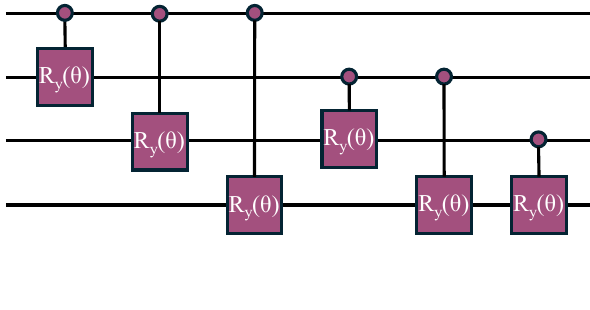}
          \vspace{-2.3em}
        \caption{Dense ent. layer}
        \label{fig:partial_layer_dense}
    \end{subfigure}
    \begin{subfigure}[t]{0.21\textwidth}
        \centering
        \includegraphics[width=\linewidth, trim=-1.25cm 3.5cm 18cm 0.3cm, clip]{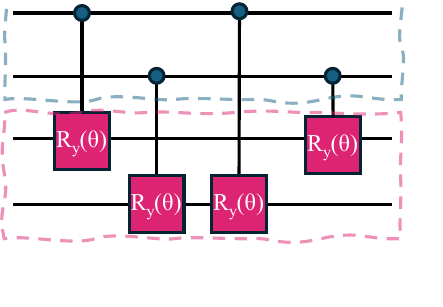}
          \vspace{-2.3em}
        \caption{Partial ent. layer}
        \label{fig:partial_layer_partial}
  
    \end{subfigure}
      \vspace{-0.5em}
    \caption{Dense entangling layer (left) and partial entangling layer (right) for $n=4$.}
    \label{fig:partial_layer}
  \vspace{-0.8em}
\end{wrapfigure}

\paragraph{Full Embedding with All Amplitudes.} 
We first describe the \textit{Full} embedding strategy in which all the possible amplitudes are used to encode data, obtaining a more expressive representation.
As a first step, the enriched input vector $\gamma(\mathbf{x})$, obtained via positional encoding, is mapped to a normalised vector $\mathcal{M}(\mathbf{x})$ of $2^n$ \textit{amplitudes}. This requires addressing three key challenges: (i) decoupling the dimensionality of $\mathbf{x}$ from the hardware-constrained number of amplitudes $2^n$; (ii) ensuring that $\mathcal{M}(\mathbf{x})$ retains sufficient representational structure for the downstream quantum model; and (iii) satisfying the normalisation constraint $\sum_{i\leq 2^n}|\alpha_i|^2=1$, where $\alpha_i$ denotes the $i$-th amplitude.
To this end, we employ a lightweight Multi-Layer Perceptron (MLP) with an output layer of size $2^n$. A ReLU activation function is applied at the end of the MLP to ensure that all amplitudes are non-negative and to increase sparsity, thereby facilitating quantum state preparation \citep{AAE}. Finally, the amplitudes are normalised to guarantee compatibility with quantum embedding requirements.
\paragraph{Dual-Branch Embedding.} 
Although recent work has improved the efficiency of amplitude state preparation \citep{fast_QAE}, encoding $2^n$ amplitudes on a quantum device remains a significant challenge \citep{QAE1, QAE2, complexityQuantumStatePrep}. Since the state preparation requires up to $\mathcal{O}(k)$ gates, where $k$ is the number of amplitudes, exploiting the full quantum space can be unfeasible on current NISQ devices \citep{NISQ_preskill}.
On the other hand, there is no intrinsic reason to treat positional and view-dependent features identically within the quantum embedding. Drawing inspiration from the original NeRF architecture, in which view-dependent features are encoded after positional features have already been processed, we introduce a \textbf{dual-branch} that separates and independently processes these components, while exponentially reducing the number of amplitudes required. 
Formally, we divide the total qubit budget into $n_p$ and $n_v$ qubits, assigned to positional and view-dependent encodings, respectively. Two MLPs are trained to produce $2^{n_p}$- and $2^{n_v}$-dimensional vectors, which are independently amplitude-encoded and then composed into a tensor product state:
\begin{equation}
|\phi(x)\rangle = |\phi_p(x_p)\rangle \otimes |\phi_v(x_v)\rangle.
\end{equation} \label{eq:DB_enc}
\hspace{-3pt}In the simplest case where $n_p = n_v = n/2$, the total number of amplitudes is reduced to $2^{n/2+1}$, offering an exponential reduction $\color{black}{\text{in } n\ \text{(e.g., }2^{n} \text{ vs.\ } 2^{n/2+1})}$ relative to the Full 3D-QISR approach. Furthermore, the dual-branch strategy also yields an exponential reduction in the number of parameters required for the MLPs. In particular, our Dual-Branch 3D-QISR model can scale up to more than $15$ qubits while maintaining fewer parameters than the classical NeRF model with approx. $590k$ parameters. Note that in our settings with $8$ qubits, Full 3D-QISR and Dual-Branch 3D-QISR have respectively $222k$ and $297k$ parameters. Amplitudes, parameters, and gates for different numbers of qubits are discussed in detail in the App.~\ref{app:scalability} (see Table \ref{tab:3D-QISR_comparison} for a summary).
\subsection{Quantum Circuit Design} 
\label{sub:quantum_design}
Following quantum embedding, the encoded data is processed by a variational quantum circuit with learnable parameters \( V(\boldsymbol{\theta}) \). 
In line with prior work on real-valued quantum feature spaces \citep{Wang2025QVFs}, we restrict the variational ansatz to the real subspace of the Hilbert space by employing only \( R_Y \) rotations. This design choice simplifies gradient-based optimisation and increases resilience to hardware noise, without significantly limiting model expressivity for the task considered. 
The circuit is composed of three elementary modules: (i) a single-qubit rotational layer consisting of \( R_Y(\theta) \) gates applied independently to each qubit; (ii) a \textit{dense entangling layer} that applies controlled-\( R_Y(\theta)_{(i,j)} \) gates across all pairs of qubits $i>j$ (Fig.~\ref{fig:partial_layer}, top); (iii) a \textit{partial entangling layer} that applies, for each $i$ in $n_p$ and for each $j$ in $n_v$, a controlled-\( R_Y(\theta)_{(i,j)} \) (Fig. \ref{fig:partial_layer}, bottom).
In the Full 3D-QISR, we build a circuit as a sequence of \( \ell \geq 1 \) blocks, each composed of a dense entangling layer followed by a rotational layer over all \( n \) qubits. A graphical representation for $\ell=3$ on $6$ qubits is given in Fig.~\ref{fig:visual_method_a}.
In the Dual-Branch 3D-QISR, we first apply a full entangling layer between the $n_p$ qubits allocated for the positional features, followed by a rotational layer over the corresponding qubits.
After these steps, the state of the system can be written as 
\begin{equation}
    (V_p(\theta) \otimes I_v) (|\phi_p(x_p)\rangle \otimes |\phi_v(x_v)\rangle) = V_p(\theta) |\phi_p(x_p)\rangle \otimes |\phi_v(x_v)\rangle
\end{equation}
where $V_p$ is the operator corresponding to the two positional layers. 
This design ensures that positional information is first internally processed and entangled, mirroring the sequential encoding strategy used in classical NeRFs.  
Next, the view-dependent amplitude embedding is introduced, and a partial dense entangling layer is applied between the \( n_p \) positional qubits and the \( n_v \) view-dependent qubits, combining the two feature spaces. A global rotational layer is then applied to all \( n \) qubits. By entangling states corresponding to positional and view-dependent features, we ensure that the model can learn the correlation between all the input coordinates.
Finally, to further enhance model expressivity, the circuit can optionally be extended with \( \ell - 2\) additional blocks, each comprising a dense entangling layer followed by a rotational layer over all \( n \) qubits. A scheme of the model for $\ell = 3$ on $6$ qubits is given in Fig.~\ref{fig:visual_method_b}.

\subsection{Parity-based Measurements}\label{ssec:parity_based_measurement}

At the end of a quantum circuit, the quantum state is typically converted into classical data via measurement. In this work, we measure the system in the computational basis, corresponding to projective measurements in the eigenbasis of the Pauli-\( Z \) operator.
To mitigate the effects of barren plateaus \citep{BP, BP_review}, we employ local measurements. Specifically, we define a family of single-qubit observables acting on individual qubits as
$\hat{O}_i = \mathbb{I}^{\otimes (i-1)} \otimes Z \otimes \mathbb{I}^{\otimes (n-i)}$ for $i = 1, \ldots, n,$
where \( Z \) denotes the Pauli-\( Z \) operator and \( \mathbb{I} \) is the identity operator on a single qubit. In practice, we write \( \hat{O}_i = Z_i \), indicating a Pauli-\( Z \) measurement on the \( i \)-th qubit.
These local projective measurements yield bitstrings corresponding to independent measurements on each qubit. Local observables are both hardware-efficient and less susceptible to barren plateaus compared to global observables, and have been shown to enhance the trainability of variational quantum algorithms \citep{BP_mitigation1, BP_mitigation2}.
The output of the circuit is given by the expectation value
$
  \mathcal{O}(\mathbf{x}) = \langle \phi_{\text{in}} | V(\boldsymbol{\theta})^\dagger \hat{O} V(\boldsymbol{\theta}) | \phi_{\text{in}} \rangle,  
$
where \( \mathbf{x} \) is the input (derived from positional and view-dependent features), \( V(\boldsymbol{\theta}) \) is the parameterised quantum circuit, and the input state \( |\phi_{\text{in}}\rangle = |\phi_{p}\rangle \otimes |\phi_{v}\rangle \) is prepared via a dual-branch amplitude embedding. The output \( \mathcal{O}(\mathbf{x}) \) is therefore a vector of \( n \) real numbers in the interval \([-1, 1]\).
To obtain outputs suitable for the considered task, we transform \( \mathcal{O}(\mathbf{x}) \) into a 4-dimensional vector representing RGB colour channels and a volumetric density. This transformation proceeds in two stages. First, for each of the four output components, we select a subset of qubits \( \mathcal{C}_i \subset \{1, \ldots, n\} \) and compute the average of their corresponding expectation values, i.e., 
$
\tilde{o}_i(\mathbf{x}) = \frac{1}{|\mathcal{C}_i|} \sum_{j \in \mathcal{C}_i} \mathcal{O}_j(\mathbf{x}),
$
where \( \mathcal{O}_j(\mathbf{x}) \) is the expectation value associated with qubit \( j \). This step, referred to as \textit{parity averaging}, aggregates information from specific qubit subsets into semantically meaningful outputs. Finally, we clip the resulting values to $[0, 1]$. 
The resulting vector \( \mathbf{o}(\mathbf{x}) = (o_1(\mathbf{x}), \ldots, o_4(\mathbf{x})) \in [0, 1]^4 \) can then be used to reconstruct a novel view as described in \citep{NERF}. The novel view is then used to compute the standard MSE loss by comparing it with the ground truth. 

\subsection{Output Scaling}\label{ssec:output_scaling}

Variational quantum circuits are known to suffer from the \textit{exponential concentration} phenomenon \citep{exponential_conc}, which refers to the tendency of their output distributions to concentrate exponentially around their mean as the number of qubits increases. This effect significantly limits the expressive power of such circuits, particularly in representing distributions with high variance. 
To mitigate the limitations introduced by exponential concentration and improve the trainability of our model, we introduce a learnable scaling factor \( \alpha_c \) applied to the output associated with each channel.
The final output $\alpha_c \mathcal{O}_c(\mathbf{x})$ is then clipped to $[0,1]$.

\begin{wrapfigure}[7]{r}{0.37\columnwidth}
\vspace{-2.15em}
  \centering
  \vspace{-0.15em} %
  \includegraphics[trim=1em 8em 28.5em 0, clip, width=0.35\columnwidth]{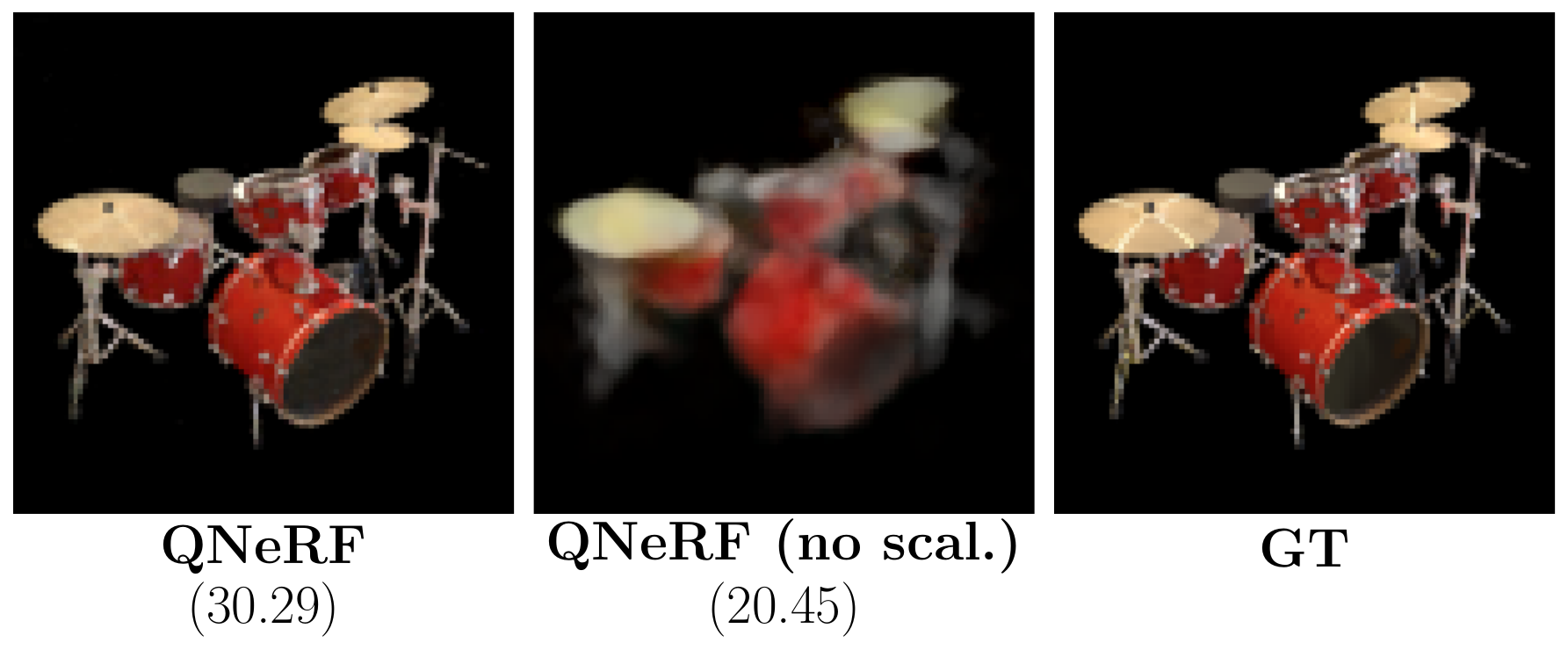}
\caption{A \textit{drums} view with (left) and without (right) output scaling.}
  \label{fig:output_scaling}
  \vspace{-0.2em} %
\end{wrapfigure}

This modification widens the range of output values, thereby counteracting the tendency of the circuit outputs to collapse around their mean, and acts in practice as a \quotes{de-concentration} layer. Empirically, we found that output scaling plays a crucial role in enhancing model performance, in particular with regard to the predicted density (see Fig.~\ref{fig:output_scaling}).

\section{Experimental Evaluation}
\label{sec:exp}

\begin{table}[bp]

\caption{Final PSNR (dB) of the proposed models compared to classical baselines for \textbf{Blender scenes (top)} and \textbf{LLFF scenes (bottom)}. Each value is averaged over 5 seeds and presented with the standard deviation. Training is stopped when the testing PSNR decreases or after 50 epochs.}
\label{tab:combined_psnr}
\centering

\resizebox{\columnwidth}{!}{
\begin{tabular}{l l l l l l}
\toprule
{\textbf{Model}} & \textit{Materials} & \textit{Ficus} & \textit{Lego} & \textit{Drums} & {Avg.} \\
\midrule
Full 3D-QISR 
 & \textbf{33.88$\pm$0.16} & \textbf{30.26$\pm$0.21} & \textbf{34.47$\pm$0.04} & \textbf{28.07$\pm$0.05} & \textbf{31.67$\pm$0.11} \\
Dual-Branch 3D-QISR 
 & 29.94$\pm$0.31 & 28.59$\pm$0.27 & 31.32$\pm$0.25 & 25.63$\pm$0.19 & 28.87$\pm$0.26 \\
\addlinespace[0.5em]
NeRF 
 & 29.90$\pm$0.19 & 29.74$\pm$0.17 & 31.79$\pm$0.16 & 26.70$\pm$0.13 & 29.53$\pm$0.16 \\
NeRF ($300k$ p.)
 & 28.08$\pm$0.57 & 29.91$\pm$0.16 & 29.97$\pm$0.61 & 26.31$\pm$0.28 & 28.57$\pm$0.22 \\
mip-NeRF &
30.53$\pm$0.69 & 29.81$\pm$0.34 & 32.73$\pm$0.27 & 27.09$\pm$0.16 & 30.04$\pm$0.37 \\
Instant NGP
 & 27.50$\pm$0.42 & 26.82$\pm$0.31 & 28.95$\pm$0.26 & 24.26$\pm$0.13 & 26.88$\pm$0.15 \\
\bottomrule
\end{tabular}%
}

\vspace{1em} %

\resizebox{\columnwidth}{!}{
\begin{tabular}{l l l l l l}
\toprule
{\textbf{Model}} & \textit{Trex} & \textit{Room} & \textit{Horns} & \textit{Fern} & {Avg.} \\
\midrule
Full 3D-QISR 
 & \textbf{22.87$\pm$1.04} & \textbf{27.94$\pm$0.45} & \textbf{23.45$\pm$0.71} & \textbf{23.21$\pm$0.41} & \textbf{24.37$\pm$0.65} \\
Dual-Branch 3D-QISR 
 & 22.03$\pm$0.42 & 26.12$\pm$0.51 & 22.02$\pm$0.48 & 22.22$\pm$0.42 & 23.10$\pm$0.46 \\
\addlinespace[0.5em]
NeRF 
 & 22.11$\pm$0.40 & 26.68$\pm$0.50 & 21.02$\pm$0.58 & 21.96$\pm$0.37 & 22.94$\pm$0.46 \\
\bottomrule
\end{tabular}%
}
\vspace{0.1em}
\end{table}

To evaluate our proposed approach, we conduct experiments on noiseless, simulated quantum hardware across multiple scenes, repeating each experiment on five different seeds. (Sec. \ref{sub:exp_blender}). These experiments are performed with the \textit{Pennylane} framework \citep{pennylane}. In addition, we evaluate the effect of noise on the proposed architecture by computing fidelity between ideal outputs and outputs obtained after transpilation on realistic hardware and noisy gate implementations and measurements (Sec.~\ref{sub:noise_resilience}). These experiments were carried out with IBM's \textit{qiskit} framework.    
To assess the effectiveness of 3D-QISR in rendering novel views, we select a representative subset of eight scenes: four from the Blender dataset, obtained from a synthetic setting, and four from the LLFF dataset \citep{llff}, following the evaluation protocol established in \citep{NERF}. All images used during training and testing are downscaled to reduce training time, as quantum simulation is computationally expensive, and currently, there are no GPU-based simulation frameworks compatible with our proposed method. In particular, training a 3D-QISR model required up to 50 hours of CPU time. Additional training details are provided in App.~\ref{app:training}.

Finally, it is important to emphasise that, as this represents the first QML-based approach for novel-view synthesis, direct comparison with other baselines is inherently challenging. To establish a meaningful, foundational comparison, we compare both Full and Dual-Branch 3D-QISR with the original classical NeRF architecture provided in \cite{NERF}. In addition, for the Blender dataset, we evaluate (i) a NeRF architecture with reduced parameter count to match our model sizes, (ii) mip-NeRF \cite{mip-NERF}, and Instant NGP \cite{instant-ngp}. 
A summary of the results is provided in Table  \ref{tab:combined_psnr}, which summarises peak signal-to-noise ratio (PSNR). We provide details and visualisations in App.~\ref{app:additional_metrics}.

The encoding sizes for positional and view-dependent features are set to 10 and 4, respectively, following the original NeRF model. In both quantum models, the encoding MLPs consist of three fully connected layers with a hidden dimension of 256, and output equal to the number of amplitudes, corresponding to $2^8$ for the Full 3D-QISR, and $2^5$ for the  Dual-Branch 3D-QISR. 
Both ansätze are constructed as described in Sec.~\ref{sub:quantum_design} with $\ell = 1$ and $\ell = 2$ for Full and Dual-Branch models, respectively. For both models, we selected the smallest meaningful value of $\ell$ (the Dual-Branch model requires at least 2 layers to include interaction between positional and view-dependent features). 
Initial values for the quantum parameters are chosen using \textit{identity initialisation} \citep{Grant2019initialization}, a common initialisation strategy to mitigate barren plateaus.
To balance model expressivity with computational complexity, we selected models with $8$ qubits.
Finally, the parity measurement assigns $2$ qubits for each output channel.
\paragraph{Blender Dataset.}
\label{sub:exp_blender}
We selected the \textit{materials}, \textit{ficus}, \textit{lego}, and \textit{drums} scenes from the Blender dataset. Each image was downscaled via average pooling to a resolution of $100{\times}100$ pixels. All models were trained on a fixed subset of 100 training images and evaluated on a fixed validation set of 200 samples. Complete quantitative results are reported in Table \ref{tab:combined_psnr} (top). 
We observe that the Full 3D-QISR model achieves higher performance on each scene. Dual-Branch has a PSNR slightly lower than NeRF and mip-NeRF, but has a higher PSNR compared to other baselines with matching parameter count.
\paragraph{LLFF Dataset.} 
\label{subs:LLFF_setting}
For the LLFF dataset, we selected the scenes \textit{horns}, \textit{fern}, \textit{trex}, and \textit{room}. Each image was downscaled to a resolution of $63{\times}47$ pixels. For each run, the set of images was split into training and testing sets in an $80$-$20$ split. Complete quantitative results are reported in Table \ref{tab:combined_psnr} (bottom).
The results on this dataset are similar to those on the Blender dataset, with minor differences. First, the average PSNR is in general lower, as the represented scenes are more complex. As a consequence, the gap between the Full 3D-QISR and NeRF performance is lower (slightly more than 1 dB). As before, the Dual-Branch model has a similar performance to the classical model, but this time it performs slightly better. The standard deviation on each experiment is higher: this depends on the fact that each seed has different splits, as, in contrast to the Blender dataset, there is no fixed split.

\paragraph{Large-Scale Evaluation on \textit{lego}.}

In addition, we evaluate our proposed Full 3D-QISR model on the
\textit{lego} scene at its original resolution of $800\times800$ pixels,
and compare its performance against NeRF and mip-NeRF. All models are
trained for $300$k steps under the same training budget.

\begin{wraptable}{r}{0.34\textwidth}
    \centering
    \vspace{-0.8em}
    \resizebox{\linewidth}{!}{%
        \begin{tabular}{l r}
            \toprule
            \textbf{Model} & \textbf{PSNR} \\
            \midrule
            Full 3D-QISR & \textbf{26.045 dB} \\
            NeRF         & 23.921 dB \\
            mip-NeRF     & 25.235 dB \\
            \bottomrule
        \end{tabular}
    }
    \caption{Average PSNR on the \textit{lego} scene at
    $800\times800$ resolution after $300$k training steps.}
    \label{tab:psnr_comparison_lego}
    \vspace{-0.8em}
\end{wraptable}

\begin{figure}[htb]
    \centering
    \includegraphics[
        width=\textwidth,
        trim=250bp 0 0 17bp,
        clip
    ]{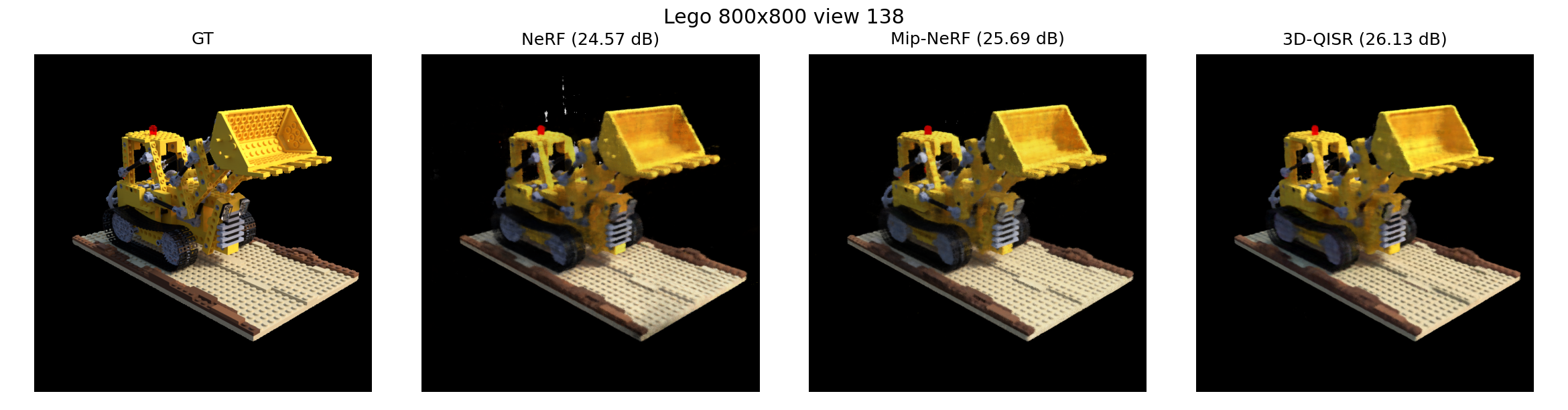}
   \caption{Qualitative comparison on an example view of the
\textit{lego} scene at $800{\times}800$ pixel resolution. The values in
parentheses report the PSNR relative to the ground truth.}
    \label{fig:lego_large_scale}
\end{figure}

While the lower-resolution experiments at $100\times100$ are trained
to convergence and already demonstrate strong reconstruction performance, this experiment investigates whether the proposed representation remains effective when scaling to substantially higher image resolutions. 
Despite none of the models having fully converged within the $300$k-step training budget, Full 3D-QISR achieves the highest PSNR, outperforming mip-NeRF by $0.81$ dB and NeRF by $2.12$ dB. These results indicate that the advantages of 3D-QISR are not limited to the low-resolution setting and suggest scalability to more demanding, high-resolution scene representations. 
Fig.~\ref{fig:lego_large_scale} provides a qualitative comparison for
an example test view. 

\subsection{Real-Hardware Noise Resilience}
\label{sub:noise_resilience} 
\begin{wrapfigure}[13]{r}{0.45\columnwidth}  
    \centering
  \vspace{-2.0em}
  \includegraphics[width=\linewidth]{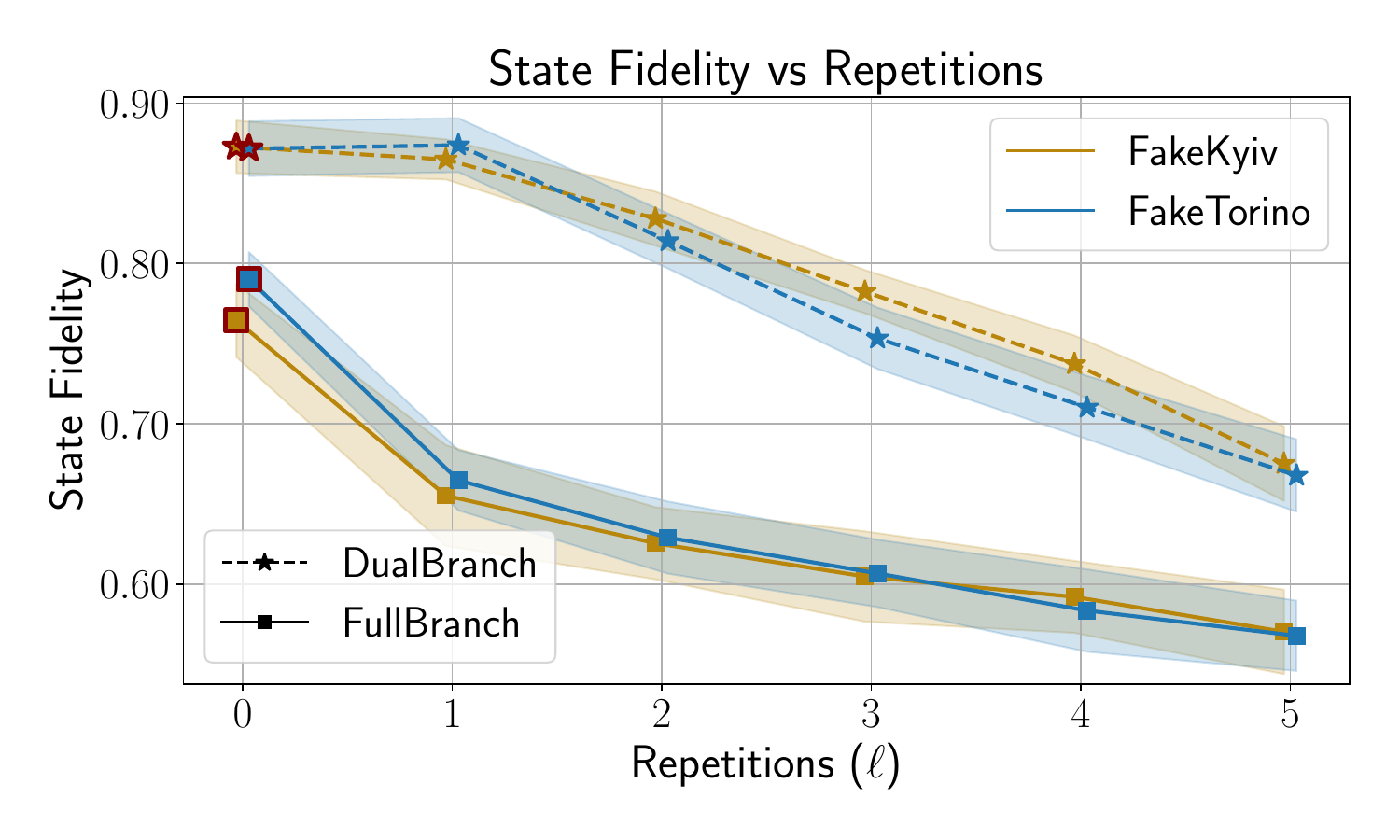}
  \caption{State fidelity for random parameter initialisations, evaluated over 50 runs for different 8-qubit ansätze and noise models.}%
  \label{fig:noise_res}
    \vspace{-0.8em}
\end{wrapfigure}

We now analyse the noise resilience of the proposed models under hardware-realistic assumptions, for different numbers of ansatz repetitions. Specifically, we employ the noise models \textit{FakeKyiv} and \textit{FakeTorino} provided by IBM \citep{qiskit_fake_provider}, which approximate the effects observed on real quantum devices. For more details and additional results, we refer to App.~\ref{app:noise}.
To quantify resilience to hardware noise, we compute the state fidelity between the noiseless circuit output and the corresponding noisy output obtained after: (i) amplitude state preparation; (ii) transpilation into the native gate set of the hardware; (iii) implementation of noisy quantum gates; (iv) noisy measurements. The results are reported in Fig.~\ref{fig:noise_res}.
The highlighted values, corresponding to $\ell=0$, represent the fidelity of the standard amplitude embedding implemented in Qiskit \citep{qiskit}. As reported in Sec.~\ref{sec:Method}, the Dual-Branch encoding requires exponentially fewer amplitudes and therefore exhibits substantially lower error rates for corresponding values of $\ell$. %

From the plot, we observe that the Dual-Branch model consistently achieves higher fidelity than the Full Branch model.
It is important to note that the fidelities reported here are severely under-optimised, as no advanced compilation or noise mitigation strategies were applied. In practical executions on real hardware, several techniques are typically employed, including optimised transpilation strategies \citep{noise_compilation, noise_compilation2}, live error mitigation methods such as Pauli Twirling \citep{Pauli_Twir, Clifford_Twir}, and post-processing approaches such as HAMMER \citep{HAMMER}. Moreover, the model may, in principle, partially adapt to systematic errors during training.
Therefore, the state fidelities shown here represent a conservative, worst-case estimate of noise resilience. Even modest optimisations in state preparation, such as approximate amplitude embedding \citep{AAE}, can already yield significant improvements in fidelity.

\subsection{\textcolor{black}{Inference Under Noise Models}}
\label{sub:inference_under_noise}
\begin{wraptable}{r}{0.5\textwidth} %
  \vspace{-1em}
  \caption{PSNR and SSIM for noise levels on the \textit{lego} scene.}
\label{tab:stacked_noise_metrics}
\centering
\small %
\setlength{\tabcolsep}{3pt} %
\begin{tabular}{l cccc cccc}
\toprule
{\textbf{Metric}} & \multicolumn{4}{c}{\textbf{Gaussian ($\sigma$)}} & \multicolumn{4}{c}{\textbf{Readout ($p$)}} \\
\cmidrule(lr){2-5} \cmidrule(lr){6-9}
 & 0 & .01 & .05 & .1 & 0 & .001 & .01 & .1 \\
\midrule
PSNR & 34.5 & 34.4 & 32.0 & 25.7 & 34.5 & 34.5 & 34.0 & 25.5 \\
SSIM & .983 & .983 & .979 & .946 & .983 & .983 & .983 & .954 \\
\bottomrule
\end{tabular}
  \vspace{-1.3em}
\end{wraptable}
We provide an evaluation of Full 3D-QISR under two ideal noise models. First, following the procedure in \citep{Wang2025QVFs}, we assume that circuit noise can be modelled as zero-mean Gaussian perturbations with varying standard deviations in the parameters of parametric gates, where higher values correspond to higher hardware noise. In particular, we evaluate the effect of Gaussian noise on the inference capabilities on the \textit{lego} scene for standard deviation $\sigma \in \{0.01, 0.05, 0.1\}$. We report in Table \ref{tab:stacked_noise_metrics} (top) both PSNR and SSIM over the seeds of the \textit{lego} dataset (see Fig.~\ref{fig:noise_combined} in App. \ref{app:additional_metrics} for qualitative results). Note that $\sigma=0.01$ produces a slight degradation of the results, with minimal change in SSIM. On the other hand, $\sigma=0.05$ produces a degradation of 2.5 dB in PSNR, with some visible visual artefact (Fig.~\ref{fig:noise_combined}, centre image).

Similarly, in Table \ref{tab:stacked_noise_metrics} (bottom) we evaluate the impact of readout error (i.e., bit flip) under the same assumptions. We provide some qualitative results in Fig.~\ref{fig:noise_combined} (bottom). In particular, we assume a probability of symmetric readout error for $p\in\{0.001, 0.01, 0.1\}$.  We observe that $p=0.001$ causes no degradation in the model performance. The value $p=0.01$, comparable to current hardware readout errors, shows a small decrease in performance. Finally, for $p=0.1$ we observe a substantial performance loss (i.e.,~the view appears darker, Fig.~\ref{fig:noise_combined} bottom right). This depends on the fact that each output is shifted closer to zero (i.e., readout error maps each output channel $o_i \mapsto (1-2p)o_i$). For this reason, we suppose output scaling can mitigate readout error during training. 

\section{Discussion and Limitations}
\label{sec:Discussion}

The experimental results indicate that the Full 3D-QISR model consistently outperforms the NeRF baseline with less than half of the parameters. The Dual-Branch model has reduced performance (comparable with the classical model), again with a reduced parameter count, but presents a much higher noise tolerance (Fig.~\ref{fig:noise_res}) due to the branched encoding and the partial entangling layer. 
One limitation of our current study is the computational cost required to perform large-scale quantum simulations. Currently, training 3D-QISR models requires substantially more time than classical baselines due to the overhead of simulating quantum circuits on a CPU. We expect this gap to shrink as both quantum simulators and hardware accelerators improve, and with the integration of GPU-accelerated quantum simulators (e.g.~\citep{GPU_sim1, GPU_sim2}). 
On the other hand, deploying these models on real-world quantum hardware may introduce challenges depending on sampling and hardware noise, limited hardware connectivity, and the complexity of gradient computation, which must be addressed. 
It is important to note that this work focuses on the fundamental challenge of introducing the gate-based QML paradigm into volumetric neural novel-view systems. While it has limitations at this exploratory stage, we believe it opens up a new research direction with many potential improvements and advances in future.
\section{Conclusion}
This work introduced two hybrid quantum-enhanced models for novel-view synthesis: {Full 3D-QISR} and {Dual-Branch 3D-QISR}. While QVFs had previously demonstrated the potential of quantum embeddings for 2D and 3D field representation learning with signal supervision of the target dimensionality, we confirm and extend their properties in a new and more challenging setting, i.e., 3D representation learning from 2D observations. 
Our experiments show that {Full 3D-QISR} shows promising results compared to classical baselines, resulting in higher PSNR on each scene considered, while using fewer than half the parameters of the strongest models. On the other hand, {Dual-Branch 3D-QISR} shows similar performance to classical baselines, with a reduced parameter count and a more scalable and noise-resistant architecture compared to Full 3D-QISR (more than $0.8$ state fidelity without any error mitigation techniques), showing promising compatibility with current or near-future quantum hardware. 
While many challenges remain, as discussed in the previous section, we believe these results represent an early but concrete step toward assessing the potential of quantum-enhanced representations for volumetric novel-view synthesis, marking a step forward toward practical quantum usefulness in computer vision. 

\bibliography{bibliography}
\bibliographystyle{unsrtnat}

\newpage
\appendix

\section*{Appendix}
We provide a brief introduction to Neural Radiance Fields (App.~\ref{app:bg_nerf}) and Quantum Computing (App.~\ref{app:bg_qc}). 
In App.~\ref{app:scalability} we discuss the scalability (in terms of amplitudes to encode and number of parameters) of the proposed quantum models. In App.~\ref{app:training} we provide some details of implementation and training.
We conduct an ablation study in App.~\ref{app:ablation_study}, in which we evaluate a ``de-quantised'' 3D-QISR model, showing that the QNN is a fundamental component of the proposed model. 
In App.~\ref{app:3dgs} we provide an additional comparison with standard 3D Gaussian Splatting on the same Blender scenes and under the same data and resolution settings, to contextualise the reconstruction quality of 3D-QISR against a state-of-the-art scene representation based on a fundamentally different rendering paradigm.
We discuss relevant metrics for novel-view rendering in App.~\ref{app:additional_metrics}, also providing some results and visualisations on the Blender dataset. In App.~\ref{app:noise}, we discuss the experimental details for the noise evaluation discussed in Sec.~\ref{sub:noise_resilience} of the main paper, and we provide additional details and statistics on the circuits transpiled on \textit{FakeKyiv} and \textit{FakeTorino}. \textcolor{black}{In App.~\ref{app:h} we provide an ablation study on the number of qubits. Then, in App.~\ref{app:ext-time} we compute the expected execution time on \textit{FakeTorino} hardware. In App.~\ref{app:j}, we evaluate the initial gradient magnitude to evaluate model trainability. Finally, in App.~\ref{app:l} and App.~\ref{app:m} we evaluate how 3D-QISR can be integrated with other tasks and models: in particular, in App.~\ref{app:l} we perform the task of Mesh Extraction for the models trained on novel-view rendering, and in \ref{app:m} we evaluate the integration of 3D-QISR with mip-NeRF \citep{mip-NERF}.}

\section{Background on NeRF}
\label{app:bg_nerf}

This section provides a brief background on the idea behind NeRF. For an in-depth introduction, we refer to \citep{survey_NeuralFieldsVisualComputing}.

\subsection{Scene Representation}

Neural Radiance Fields model a 3D scene as a continuous volumetric function that maps spatial locations and viewing directions to emitted radiance and volume density. Formally, a scene is represented by a function 
\begin{equation}
    F : (\mathbf{x}, \mathbf{d}) \mapsto (\mathbf{c}, \sigma),
\end{equation}

where $\mathbf{x} = (x, y, z) \in \mathbb{R}^3$ denotes the spatial coordinate, $\mathbf{d} = (\theta, \phi)$ represents the viewing direction, $\mathbf{c} = (r, g, b) \in [0,1]^3$ denotes the RGB colour at that point along the viewing ray, and $\sigma \in \mathbb{R}_{\geq 0}$ represents the volume density, corresponding to the opacity at that location.

In practice, this function is approximated using a multilayer perceptron, or more generally, a neural network, trained to fit a sparse set of 2D images of a scene by minimising a photometric reconstruction loss.

\subsection{Volume Rendering}

The colour of a pixel in an image is synthesised by integrating the radiance accumulated along a camera ray as it traverses the 3D scene. Given a ray parameterised as $\mathbf{r}(t) = \mathbf{o} + t\mathbf{d}$, where $\mathbf{o}$ is the ray origin (i.e., the camera centre) and $\mathbf{d}$ is the viewing direction (corresponding to a unit vector in $\mathbb{R}^3$), the colour $C(\mathbf{r})$ of the ray is computed using the volume rendering equation:

\begin{equation}
C(\mathbf{r}) = \int_{t_n}^{t_f} T(t)\, \sigma(\mathbf{r}(t))\, \mathbf{c}(\mathbf{r}(t), \mathbf{d})\, dt,
\end{equation}
where $\sigma(\mathbf{r}(t))$ denotes the volume density at point $\mathbf{r}(t)$, $\mathbf{c}(\mathbf{r}(t), \mathbf{d})$ is the emitted RGB colour in direction $\mathbf{d}$, and $[t_n, t_f]$ are the near and far bounds of the ray.

The term $T(t)$ represents the accumulated transmittance, that is, the probability that a photon travels from $t_n$ to $t$ without being absorbed, and is defined as:

\[
T(t) = \exp\left( -\int_{t_n}^{t} \sigma(\mathbf{r}(s))\, ds \right).
\]

In practice, the continuous integral is approximated numerically using a discrete sampling scheme. The ray is divided into $N$ stratified or uniform intervals $\{t_i\}_{i=1}^N$, and the colour is estimated using quadrature:

\[
\hat{C}(\mathbf{r}) = \sum_{i=1}^{N} T_i \left(1 - \exp(-\sigma_i \delta_i)\right) \mathbf{c}_i,
\]

where $\sigma_i = \sigma(\mathbf{r}(t_i))$, $\mathbf{c}_i = \mathbf{c}(\mathbf{r}(t_i), \mathbf{d})$, $\delta_i = t_{i+1} - t_i$ is the distance between adjacent samples, and $T_i = \exp\left(-\sum_{j=1}^{i-1} \sigma_j \delta_j\right)$ denotes the discrete approximation of the transmittance.

Finally, this rendering formulation enables gradient-based optimisation of the scene representation by comparing synthesised views with ground-truth images, making it possible to reconstruct detailed volumetric radiance fields from sparse observations.

\section{Background on Quantum Computing}
\label{app:bg_qc}

This section presents some key concepts of quantum computing. For a detailed introduction, we refer to \citep{intro_to_QC}.

A \textit{qubit} is a two-level quantum system described by a complex-valued unit vector in a 2-dimensional Hilbert space:
\[
|\psi\rangle = \alpha|0\rangle + \beta|1\rangle, \quad \text{where } \alpha, \beta \in \mathbb{C}, \quad |\alpha|^2 + |\beta|^2 = 1,
\]
and $|0\rangle$, $|1\rangle$ represent the quantum analogue of classical bit values $0$ and $1$.

Similarly, an $n$-qubit system can be represented by a $2^n$-dimensional Hilbert space:
\[
|\psi\rangle = \sum_{i=0}^{2^n - 1} \alpha_i |i\rangle, \quad \text{where } \sum_i |\alpha_i|^2 = 1.
\]
Given two quantum states $|\phi\rangle$ and $|\psi\rangle$ of $n$ and $m$ qubits, respectively, the state $|\phi\rangle\otimes|\psi\rangle$, where $\otimes$ represents the tensor product, can be represented in a quantum system of size $n+m$. 

\subsection{Quantum Gates and Circuits}

Quantum computations are realised via the application of unitary transformations to quantum states. A unitary operator $U \in \mathbb{C}^{2^n \times 2^n}$ evolves a quantum state $|\psi\rangle$ to a new state $|\psi'\rangle=U|\psi\rangle$. Physically, such operations are implemented as \emph{quantum circuits}, which can be decomposed into a sequence of elementary operations known as \emph{quantum gates}.

In practice, most hardware platforms support only one- and two-qubit gates. Nevertheless, these gates form a universal set, meaning they can be composed to approximate any arbitrary unitary operation.

A quantum gate is called \emph{parametric} if its action depends on a continuous parameter. An example is the single-qubit rotation around the $Y$ axis:
\[
R_Y(\theta) = \begin{bmatrix}
\cos(\theta/2) & -\sin(\theta/2) \\
\sin(\theta/2) & \cos(\theta/2)
\end{bmatrix}.
\]

\subsection{Amplitude Encoding}

Given a normalised real-valued vector $\mathbf{x} \in \mathbb{R}^{2^n}$ such that $\|\mathbf{x}\|_2 = 1$, \emph{amplitude encoding} maps it into a quantum state as follows:
\[
|\phi(\mathbf{x})\rangle = \sum_{i=0}^{2^n - 1} x_i |i\rangle.
\]
This encoding technique allows for representing $2^n$ classical values using only $n$ qubits, providing an exponential compression of input data, but requires $\mathcal{O}(2^n)$ gates in the general case \citep{complexityQuantumStatePrep}. 

\section{3D-QISR Scalability: Parameters and Amplitudes}\label{app:scalability}

We analyse the scalability of the proposed Full 3D-QISR and Dual-Branch 3D-QISR models by examining how their complexity increases with the number of qubits. Specifically, we compare the number of amplitudes, classical parameters, and quantum gates required by each architecture for increasing values of $n$. A summary of these results is presented in Table~\ref{tab:3D-QISR_comparison}.  

As the number of qubits increases, the complexity of the Full 3D-QISR model grows exponentially not only in terms of the number of output amplitudes ($2^n$), but also in the number of classical parameters. This is because the encoding MLP must produce an output vector of size $2^n$, one amplitude per computational basis state. In our experimental setup, which uses a 3-layer MLP with a hidden dimension of 256, using up to 10 qubits results in fewer total parameters than those found in a standard classical NeRF model.
In contrast, the Dual-Branch 3D-QISR architecture exhibits a more favourable scaling pattern: it only requires $2^{n/2}$ amplitudes per branch. Although it uses two encoding MLPs---leading to a larger parameter count for small $n$---this approach allows the model to scale up to 18 qubits while maintaining a total parameter count comparable to that of classical NeRF. Importantly, the number of amplitudes remains within the feasibility range for implementation on near-term quantum devices. 
A critical factor in practical scalability is the preparation of the initial quantum state via amplitude encoding. In the general case, the number of quantum gates required for this preparation scales linearly with the number of amplitudes \citep{complexityQuantumStatePrep}. Under this assumption, the Dual-Branch 3D-QISR can be feasibly implemented on NISQ devices with up to 12 qubits per branch.

Recent work \citep{fast_QAE} suggests that under certain structural assumptions, the cost of amplitude state preparation can be significantly reduced. An additional promising direction is Approximate Amplitude Embedding \citep{AAE}, in which a parameterised quantum circuit is trained to approximate the desired initial state. This approach has already shown promising results in a quantum computer vision task \citep{image_classification_AE} and may also benefit quantum NeRF architectures.

Another relevant factor is the number of variational parameters $\mathbf{\theta}$ in the quantum ansatz, as this directly influences the number of circuit evaluations required to estimate gradients during training, typically via the parameter-shift rule \citep{parameter_shift_rule}. In our ansatz, each gate is associated with a single trainable parameter, making the total number of gates equal to the number of variational parameters. This number scales quadratically with the number of qubits, but in our setup it remains below 100, making the circuit suitable for current quantum hardware. Nonetheless, the adoption of gradient-free optimisation methods such as COBYLA \citep{COBYLA} or SPSA \citep{SPSA} could further reduce the shot complexity during training.

\begin{figure}[htbp]
    \centering
    \includegraphics[width=0.7\textwidth]{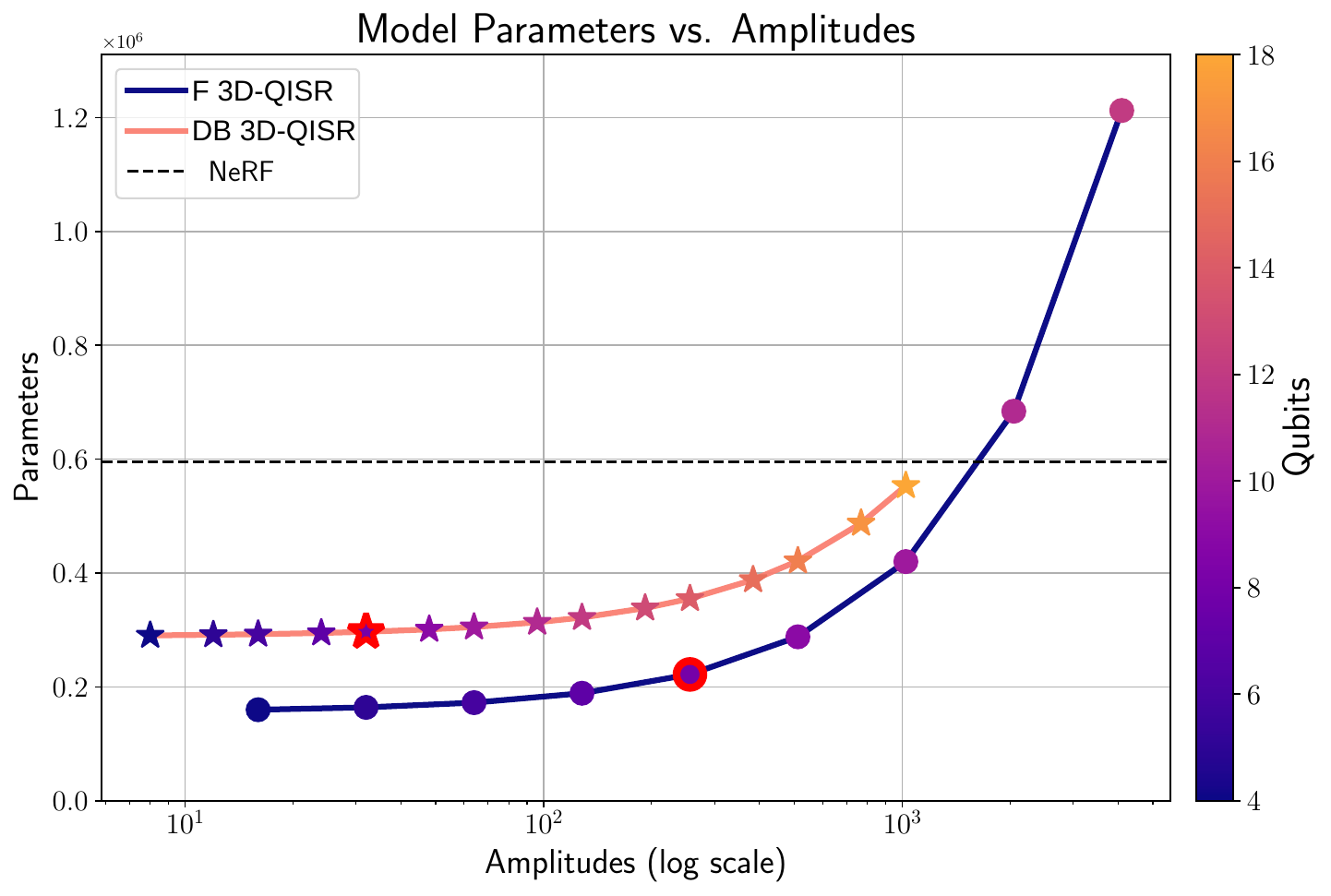}
    \caption{Comparison of Parameters vs Amplitudes for 3D-QISR models, where the Full is represented as a circle, and the Dual-Branch as a star. The points highlighted in red correspond to the values for $8$ qubits, as used in the experimental section. }
    \label{fig:your_label}
\end{figure}

\begin{table*}[htbp]
\centering
\begin{tabular}{l r r l r r l}
\toprule
\textbf{Qubits} & \multicolumn{3}{c}{\textbf{Full 3D-QISR}} & \multicolumn{3}{c}{\textbf{Dual-Branch 3D-QISR}} \\
\cmidrule(lr){2-4} \cmidrule(lr){5-7}
 & Amplitudes & Parameters ($k$) & Gates & Amplitudes & Parameters ($k$) & Gates \\
\midrule
4  & 16   & 160  & 10     & 8    & 290   & 21 \\
6  & 64   & 172   & 21     & 16   & 293   & 42 \\
\textbf{8}  & \textbf{256}  & \textbf{222}   & \textbf{36}     & \textbf{32}  & \textbf{297}   & \textbf{70} \\
10 & 1\,024 & 420 & 55     & 64   & 305   & 105 \\
12 & 4\,096 & 1\,213 & 78  & 128  & 322   & 147 \\
\bottomrule
\end{tabular}
\caption{Comparison of Full and Dual-Branch models in terms of amplitude and parameter count as a function of the number of qubits. 
The number of parameters depends on the number of qubits and is computed for an MLP of $3$ layers with hidden dimension $h=256$. 
The gate counts depend on the number of repetitions $\ell$, and are reported for $\ell = 1$ and $\ell = 2$ for Full and Dual-Branch, respectively. 
The number of qubits $n=8$ (in bold) corresponds to the value used in Sec.~\ref{sec:exp}. \textcolor{black}{Note that the number of required amplitudes in the Full model scales quadratically compared to the DB model}.}
\label{tab:3D-QISR_comparison}
\end{table*}

\section{Training Details}
\label{app:training}

We trained all models using the Adam optimiser \citep{ADAM} with an initial learning rate of $5 \times 10^{-4}$ for all parameters but the ones for the output scaling layer, which were assigned a higher learning rate of $0.01$. The learning rate is decreased with a multi-step scheduler up to $6.25\times10^{-5}$ ($1.25\times10^{-4}$ for the scaling layer). We used a batch size of 64.

Quantum circuits were simulated on CPU using the \textit{PennyLane} framework \citep{pennylane} (version 0.37), employing the noiseless \textit{default.qubit} backend. 
Each model was trained for up to $50$ epochs and tested every five epochs. 
Training is stopped if the testing PSNR starts decreasing. As can be observed in Table \ref{tab:app_stopping_epcoh} and \ref{tab:app_stopping_epcoh_llff}, all models reached convergence, apart from Full 3D-QISR on \textit{Lego} and \textit{Drums} scenes. 

\begin{table}[ht]
\caption{Summary of average stopping epochs for different models and datasets.}
\centering
\begin{subtable}[t]{0.99\linewidth}
\centering
\begin{tabular}{lcccc}
\toprule
\textbf{Model} & \textit{Materials} & \textit{Ficus} & \textit{Lego} & \textit{Drums} \\
\midrule
Full 3D-QISR & 40 & 43 & 50+ & 50+ \\
DB 3D-QISR & 34 & 24 & 36 & 18 \\
\midrule
Baseline & 16 & 19 & 24 & 19 \\
\bottomrule
\end{tabular}
\caption{Average stopping epoch across synthetic scenes.}
\label{tab:app_stopping_epcoh}
\end{subtable}

\vspace{0.5cm}

\begin{subtable}[t]{0.99\linewidth}
\centering
\begin{tabular}{lcccc}
\toprule
\textbf{Model} & \textit{Trex} & \textit{Room} & \textit{Horns} & \textit{Fern} \\
\midrule
Full 3D-QISR & 28 & 39 & 42 & 42 \\
DB 3D-QISR & 23 & 42 & 43 & 20 \\
\midrule
Baseline & 18 & 35 & 30 & 20 \\
\bottomrule
\end{tabular}
\caption{Average stopping epoch across LLFF scenes.}
\label{tab:app_stopping_epcoh_llff}
\end{subtable}
\end{table}

\begin{table}[ht]
\caption{Average training time (hours) to reach convergence on each synthetic scene.}
\label{tab:app:time_s}
\centering
\begin{tabular}{lcccc}
\toprule
\textbf{Model} & \textit{Materials} & \textit{Ficus} & \textit{Lego} & \textit{Drums} \\
\midrule
Full 3D-QISR & 39.36 & 42.31 & 49.20 & 49.20 \\
DB 3D-QISR & 48.96 & 34.56 & 51.84 & 25.92 \\
\bottomrule
\end{tabular}

\end{table}

\begin{table}[ht]
\caption{Average training time (hours) to reach convergence on LLFF datasets.}
\label{tab:app:time_b}
\centering
\begin{tabular}{lcccc}
\toprule
\textbf{Model} & \textit{Trex} & \textit{Room} & \textit{Horns} & \textit{Fern} \\
\midrule
Full 3D-QISR & 3.41 & 4.06 & 5.71 & 2.02 \\
DB 3D-QISR & 4.09 & 5.54 & 8.60 & 1.20 \\
\bottomrule
\end{tabular}

\end{table}

In Tables \ref{tab:app:time_s} and \ref{tab:app:time_b}, we report the CPU time required for each model to reach convergence. GPU-based quantum simulation was not feasible due to current limitations in PennyLane's amplitude embedding. Model components were implemented in \textit{PyTorch} (2.6) \citep{pytorch}.  We expect that the integration with GPU-accelerated quantum simulators (e.g. \citep{GPU_sim1, GPU_sim2}) could significantly reduce training time, potentially making simulated QML models more efficient than classical counterparts.

It is interesting to note that the time required by the Dual-Branch 3D-QISR is approximately $40\%$ more than the Full 3D-QISR model. This depends mostly on \textit{PennyLane} framework, as the branched amplitude embedding is not currently supported (in our work, we prepared the initial state with two subsequent partial amplitude embeddings).

\section{Ablation Study: ``Classical 3D-QISR''}
\label{app:ablation_study}

As a simple ablation study, we want to evaluate whether the increase in the performance of the proposed Full 3D-QISR model depends on the MLP feature encoding strategy or on the usage of the QNN. 
To do so, we replace the QNN of the model described in Fig.~$\ref{fig:visual_method_a}$ with an MLP with three layers and a hidden dimension of $256$, similar to the dimensions used in the MLPs in the quantum models and in classical NeRF. 
In particular, the first MLP to learn the encoding remains unchanged, i.e., ReLU plus amplitude normalisations. However, in contrast to what is done for the 3D-QISR model, where the $2^n$ amplitudes are encoded in $n$ qubits, we process the $2^n$ ``classical amplitudes'' with another MLP, resulting in a ``classical 3D-QISR". 
In more detail, the architecture of the second MLP consists of layers with dimensions $(2^n \times 256)$, $(256\times 256)$, and $(256 \times 4)$ to obtain the RGB and $\sigma$ values at the end.
\begin{table*}[ht]
\caption{Average PSNR over 5 seeds on the Blender dataset, with the addition of the ``Classical 3D-QISR'' model (last row).}
\label{tab:app:additional}
\centering
\begin{tabular}{l c c c c c}
\toprule
\textbf{Model} & \textit{Materials} & \textit{Ficus} & \textit{Lego} & \textit{Drums} & {Average} \\
\midrule
Full 3D-QISR & \textbf{33.88$\pm$0.16} & \textbf{30.26$\pm$0.21} & \textbf{34.47$\pm$0.04} & \textbf{28.07$\pm$0.05} & \textbf{31.67$\pm$0.11} \\
DB 3D-QISR & 29.94$\pm$0.31 & 28.59$\pm$0.27 & 31.32$\pm$0.25 & 25.63$\pm$0.19 & 28.87$\pm$0.26 \\
\midrule
NeRF & 29.90$\pm$0.19 & 29.74$\pm$0.17 & 31.79$\pm$0.16 & 26.70$\pm$0.13 & 29.53$\pm$0.16 \\
\midrule 
``Class. 3D-QISR'' & 23.11$\pm$0.42 & 22.60$\pm$0.06 & 21.16$\pm$0.06 & 19.03$\pm$0.05 & 21.47$\pm$1.58 \\
\bottomrule
\end{tabular}

\end{table*}

\begin{table}[ht]
\caption{Model parameters for proposed quantum models, NeRF, and ``Classical 3D-QISR''.}
\label{tab:app:parameter count}
\centering
\begin{tabular}{l c}
\toprule
\textbf{Model} & Parameters ($k$) \\
\midrule
Full 3D-QISR & 222 \\
Dual-Branch 3D-QISR & 297 \\
NeRF & 590 \\
``Classical 3D-QISR'' & 352 \\
\bottomrule
\end{tabular}
\end{table}
As reported in Table \ref{tab:app:additional}, the 3D-QISR model without the quantum component performs poorly, resulting in an average PSNR that is 8 dB lower than the standard NeRF model. It is important to note that the performance drop does not depend directly on the number of parameters: as observed in Table \ref{tab:app:parameter count}, the ``Classical 3D-QISR'' has fewer parameters than the standard NeRF model, but more than both Full 3D-QISR and Dual-Branch 3D-QISR.

This shows that the quantum neural network in the proposed architecture is fundamental for the high performance of the model.

\section{Comparison with 3D Gaussian Splatting}
\label{app:3dgs}

For broader context, we additionally compare 3D-QISR with standard 3D Gaussian Splatting (3DGS)~\citep{3DGS} on the Blender scenes. We use the same 100 training views and $100{\times}100$ image resolution adopted in our main experiments, while otherwise retaining the default optimisation configuration of the official 3DGS implementation, without scene-specific hyperparameter tuning. In particular, we use 30k optimisation iterations, spherical harmonics of degree 3, the standard densification schedule, and no antialiasing.
For the synthetic Blender scenes, 3DGS is initialised from 100k random points, following the original protocol~\citep{3DGS}. 
We stress that 3DGS belongs to a different class of scene representations: unlike 3D-QISR and the NeRF-based baselines considered in the main paper, it represents the scene explicitly through an adaptive collection of 3D Gaussians and employs splatting-based rendering rather than a continuous neural radiance field. We therefore report this comparison as broader context for reconstruction quality, rather than as a direct comparison of model complexity or computational efficiency. 

\begin{table}[ht]
\caption{PSNR (dB) comparison with 3D Gaussian Splatting on the Blender scenes. Results are averaged over five seeds and reported with standard deviation.}
\label{tab:app:3dgs}
\centering
\begin{tabular}{lccccc}
\toprule
\textbf{Model} & \textit{Materials} & \textit{Ficus} & \textit{Lego} & \textit{Drums} & \textbf{Avg.} \\
\midrule
Full 3D-QISR
& \textbf{33.88$\pm$0.16}
& 30.26$\pm$0.21
& \textbf{34.47$\pm$0.04}
& 28.07$\pm$0.05
& \textbf{31.67$\pm$0.11} \\
Dual-Branch 3D-QISR
& 29.94$\pm$0.31
& 28.59$\pm$0.27
& 31.32$\pm$0.25
& 25.63$\pm$0.19
& 28.87$\pm$0.26 \\
3DGS
& 32.29$\pm$0.02
& \textbf{30.99$\pm$0.03}
& 34.26$\pm$0.02
& \textbf{28.29$\pm$0.03}
& 31.46$\pm$0.03 \\
\bottomrule
\end{tabular}
\end{table}

Full 3D-QISR achieves reconstruction quality comparable to 3DGS overall, with a slightly higher average PSNR ($+0.21$ dB). Its performance is higher on \textit{Materials} ($+1.59$ dB) and \textit{Lego} ($+0.21$ dB), and lower on \textit{Ficus} ($-0.73$ dB) and \textit{Drums} ($-0.22$ dB). Dual-Branch 3D-QISR is consistently below 3DGS, with an average gap of $2.59$ dB. Given the fundamentally different representations and rendering pipelines, these results should be interpreted as broader context on reconstruction quality rather than as a direct efficiency comparison.

\section{Additional Metrics and Renders}
\label{app:additional_metrics}

In the main paper, we report quantitative results in terms of Peak Signal-to-Noise Ratio (PSNR), as it is the most informative metric for evaluating pixel-wise reconstruction quality in novel view synthesis tasks. PSNR is defined as
\[
    \text{PSNR}(x, \hat{x}) = 10 \cdot \log_{10} \left( \frac{L^2}{\text{MSE}(x, \hat{x})} \right),
\]
where $L$ is the maximum possible pixel value of the image (e.g., $L=1$ for normalised images), $x$ is the ground-truth image, $\hat{x}$ is the reconstructed image, and $\text{MSE}$ denotes the mean squared error.

Another relevant metric is the Structural Similarity Index (SSIM), which better correlates with human perception of image quality. SSIM between two images $x$ and $\hat{x}$ is defined as
\[
    \text{SSIM}(x, \hat{x}) = \frac{(2\mu_x \mu_{\hat{x}} + C_1)(2\sigma_{x\hat{x}} + C_2)}{(\mu_x^2 + \mu_{\hat{x}}^2 + C_1)(\sigma_x^2 + \sigma_{\hat{x}}^2 + C_2)},
\]
where $\mu_x$ and $\mu_{\hat{x}}$ denote the mean pixel intensities, $\sigma_x^2$ and $\sigma_{\hat{x}}^2$ the variances, and $\sigma_{x\hat{x}}$ the covariance between $x$ and $\hat{x}$; $C_1, C_2$ are small constants for numerical stability.

We report SSIM values for the Blender dataset in Table~\ref{tab:ssim_synthetic}, using the same experimental setting as for PSNR. We observe that the trends across models are consistent with those obtained with PSNR, confirming that our proposed approach achieves superior perceptual reconstruction quality.

\begin{figure*}[htbp]
    \centering
    \includegraphics[width=0.99\textwidth]{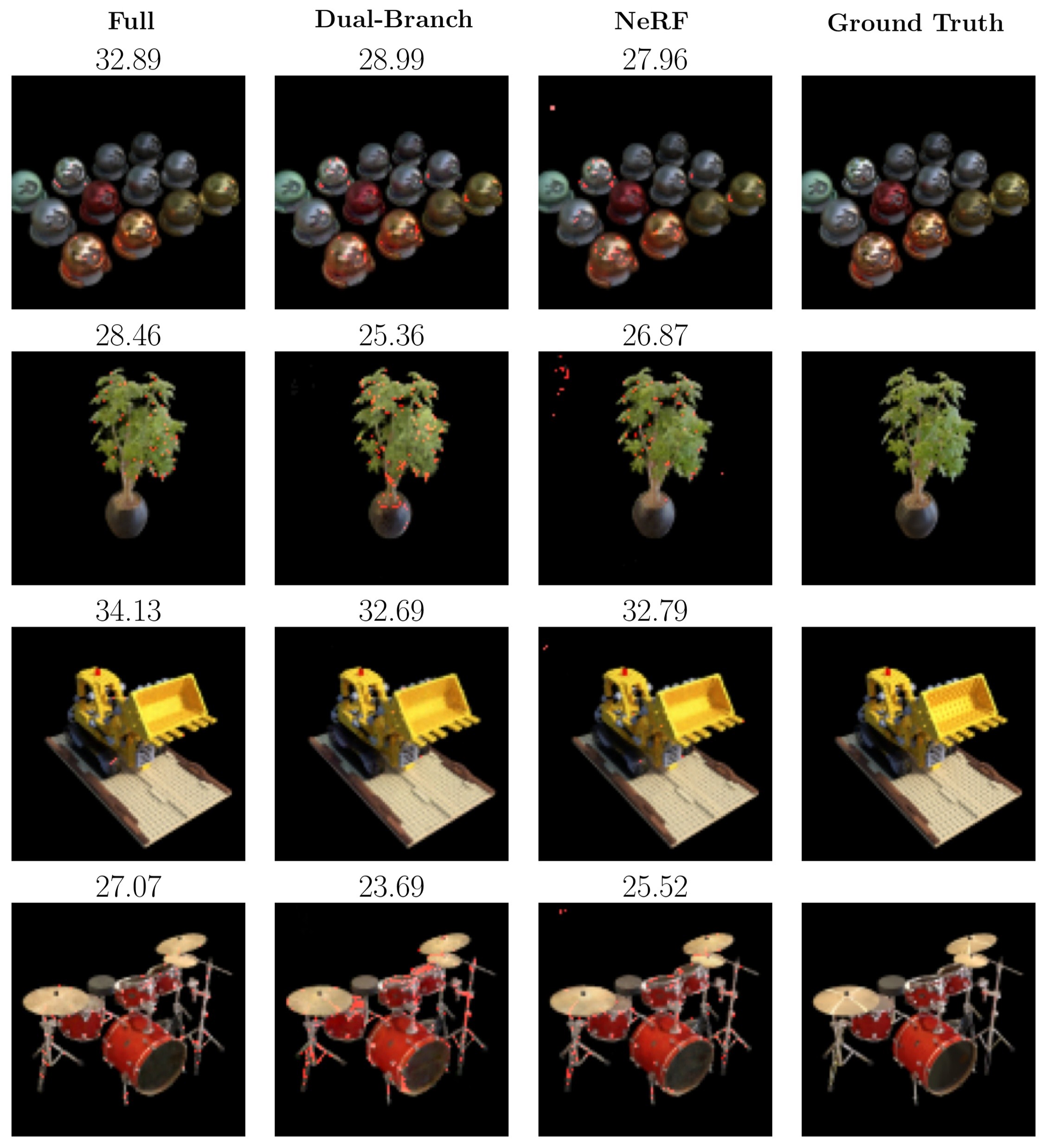}
    \caption{Visualisation of scenes from the Blender dataset \citep{NERF} (PSNR with respect to the ground truth on the top). Pixels with higher error values are highlighted in red.}  
    \label{fig:app:additional_renders}
\end{figure*}

In addition, we provide qualitative results in Fig.~\ref{fig:app:additional_renders}, showing example renders from the Blender dataset. Pixels with higher reconstruction errors are highlighted in red, providing further insights into the regions where differences among the models are most pronounced.

\begin{table*}[ht]
\caption{SSIM on the Blender dataset, under the same experimental setting given in Sec.~\ref{sec:exp}.}
\label{tab:ssim_synthetic}
\centering
\begin{tabular}{l c c c c c}
\toprule
\textbf{Model} & \textit{Materials} & \textit{Ficus} & \textit{Lego} & \textit{Drums} & {Average} \\
\midrule
Full 3D-QISR & \textbf{0.9834$\pm$0.001} & \textbf{0.967$\pm$0.001} & \textbf{0.984$\pm$0.000} & \textbf{0.955$\pm$0.001} & \textbf{0.972$\pm$0.001} \\
DB 3D-QISR & 0.964$\pm$ 0.002&0.954$\pm$0.001 &0.966$\pm$0.002 & 0.926 $\pm$0.003 & 0.953$\pm$0.016 \\
\midrule
Class. Baseline & 0.964$\pm$0.002 & 0.955$\pm$0.004 & 0.973$\pm$0.002 & 0.939$\pm$0.003 & 0.958$\pm$0.012 \\
\bottomrule
\end{tabular}

\end{table*}

\begin{figure*}[htbp]
    \centering
    \includegraphics[width=0.99\textwidth]{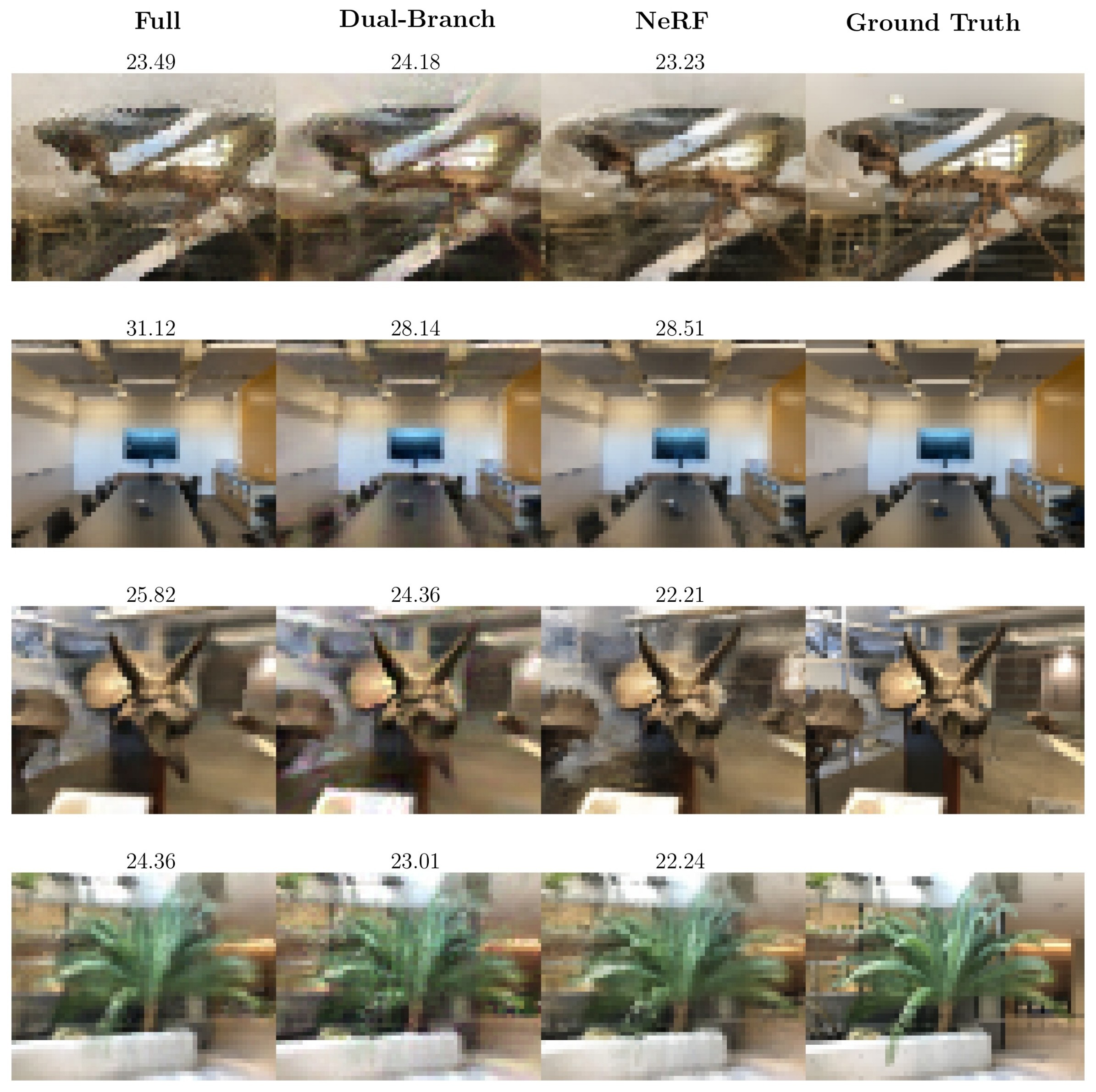}
    \caption{Visualisation of scenes from the LLFF dataset \citep{llff}. }
    \label{fig:app:additional_renders2}
\end{figure*}

\begin{figure}
    \centering
    \begin{minipage}[t]{0.31\textwidth}
        \centering
        \includegraphics[width=\textwidth]{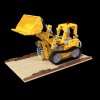}
        \subcaption*{GN ($\sigma = 0.01$).}
    \end{minipage}
    \hfill
    \begin{minipage}[t]{0.31\textwidth}
        \centering
        \includegraphics[width=\textwidth]{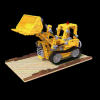}
        \subcaption*{GN ($\sigma = 0.05$).}
    \end{minipage}
    \hfill
    \begin{minipage}[t]{0.31\textwidth}
        \centering
        \includegraphics[width=\textwidth]{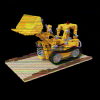}
        \subcaption*{GN ($\sigma = 0.1$.)}
    \end{minipage}

    \begin{minipage}[t]{0.31\textwidth} %
        \centering
        \includegraphics[width=\textwidth]{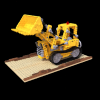}
        \subcaption*{RE ($p = 0.001$).}
    \end{minipage}
    \hfill
    \begin{minipage}[t]{0.31\textwidth}
        \centering
        \includegraphics[width=\textwidth]{001.png}
        \subcaption*{RE ($p = 0.01$).}
    \end{minipage}
    \hfill
    \begin{minipage}[t]{0.31\textwidth}
        \centering
        \includegraphics[width=\textwidth]{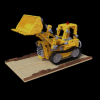}
        \subcaption*{RE ($p = 0.1$).}
    \end{minipage}
    \caption{\textcolor{black}{Novel views of the Full 3D-QISR model trained on the \textit{lego} scene, under Gaussian Noise (GN) perturbations, top row, and symmetric readout errors (RE), bottom row.}}
    \label{fig:noise_combined}
\end{figure}

Finally, in Fig.~\ref{fig:noise_combined} we provide a visual representation of the effect of Gaussian noise with different standard deviations $\sigma$ (top row), and readout symmetric error with different probabilities $p$ (bottom row).

\section{Noise Resilience: Experimental Setup and Extended Analysis}
\label{app:noise}

We provide the detailed methodology and additional results underpinning the noise resilience analysis presented in Sec.~\ref{sec:exp}. The purpose of these experiments is to assess how the different ansatz architectures respond to realistic hardware noise.

To emulate the effect of noise on real quantum hardware, we employed IBM’s publicly available noise models \texttt{FakeKyiv} and \texttt{FakeTorino} on \textit{qiskit}\citep{qiskit}. These models incorporate gate-dependent depolarisation, readout errors, and device-specific connectivity constraints. By applying them at the simulation level, we are able to obtain reproducible fidelity estimates under consistent noise conditions.
Fidelity is a standard measure of similarity between two quantum states. For two density matrices $\rho$ and $\sigma$, it is defined as:
\[
F(\rho, \sigma) = \left(\mathrm{Tr}\sqrt{\sqrt{\rho}\, \sigma \, \sqrt{\rho}}\right)^2,
\]
where $F(\rho, \sigma) \in [0,1]$, with $F=1$ indicating identical states and $F=0$ corresponding to orthogonal states. 

In particular, for pure states,
\[
\rho = |\psi_\rho\rangle \langle \psi_\rho| \quad \text{and} \quad \sigma = |\psi_\sigma\rangle \langle \psi_\sigma|,
\] 
the fidelity reduces to
\[
F(\rho, \sigma) = |\langle \psi_\rho | \psi_\sigma \rangle|^2.
\]

This metric allows us to quantitatively assess how closely the noisy quantum state $\rho_{\text{noisy}}$ approximates the ideal noiseless state $\rho_{\text{ideal}}$ after circuit execution.

In our experiments, all amplitude embeddings employ Qiskit’s \texttt{initialize} routine, followed by decomposition into the device’s native gate set with four rounds of operator expansion (\texttt{decompose(reps=4)}) to decompose the gates into the set of gates that are feasible for the hardware under consideration. This ensures that the reported fidelities account for the practical cost of state preparation.

Also, other gates must be decomposed into gates from the base set of gates of the hardware. In particular:
\begin{itemize}
    \item \textit{FakeKyiv} requires the circuit to be decomposed into \texttt{sx}, \texttt{rz}, \texttt{ecr}, and \texttt{x} gates, corresponding to $\sqrt X, R_z$, the echoed cross-resonance (ECR), and $X$ gates.
    \item \textit{FakeTorino} requires gates \texttt{sx}, \texttt{rz}, \texttt{cz}, and \texttt{x} gates, corresponding to $\sqrt X, R_z$, controlled-$Z$, and $X$ gates.
\end{itemize}

For each configuration, we evaluated the circuit output under both noiseless and noisy conditions. The primary metric is the \emph{state fidelity} between the noiseless state $\rho_{\text{ideal}}$ and the noisy state $\rho_{\text{noisy}}$ obtained after amplitude embedding, transpilation, and execution under the noise model. To mitigate bias from specific parameter values, we repeated the experiment $50$ times with randomly drawn parameter initialisations and report aggregated fidelity statistics in Fig.~\ref{fig:noise_res}.

The key finding is that the Dual-Branch ansatz achieves systematically higher fidelities compared to the Full-Branch model at equal depth. This advantage originates from its reduced reliance on exponentially large amplitude vectors and its more localised embedding strategy, which leads to a lower effective error rate.

The baseline case $\ell=0$, corresponding to the fidelity of the encoding, is highlighted in Fig.~\ref{fig:noise_res}. For the Full model, this corresponds to amplitude embedding on all qubits, while for the Dual-Branch model, it corresponds to separate embeddings on each half of the register. The intermediate case $\ell=1$ in the Dual-Branch configuration is reported for completeness but is not representative of the architecture adopted in this work, as it lacks the full view-dependent interaction layer.

\begin{figure}[htbp]
    \centering
    \begin{subfigure}[t]{0.497\textwidth}
        \centering
        \includegraphics[width=\linewidth]{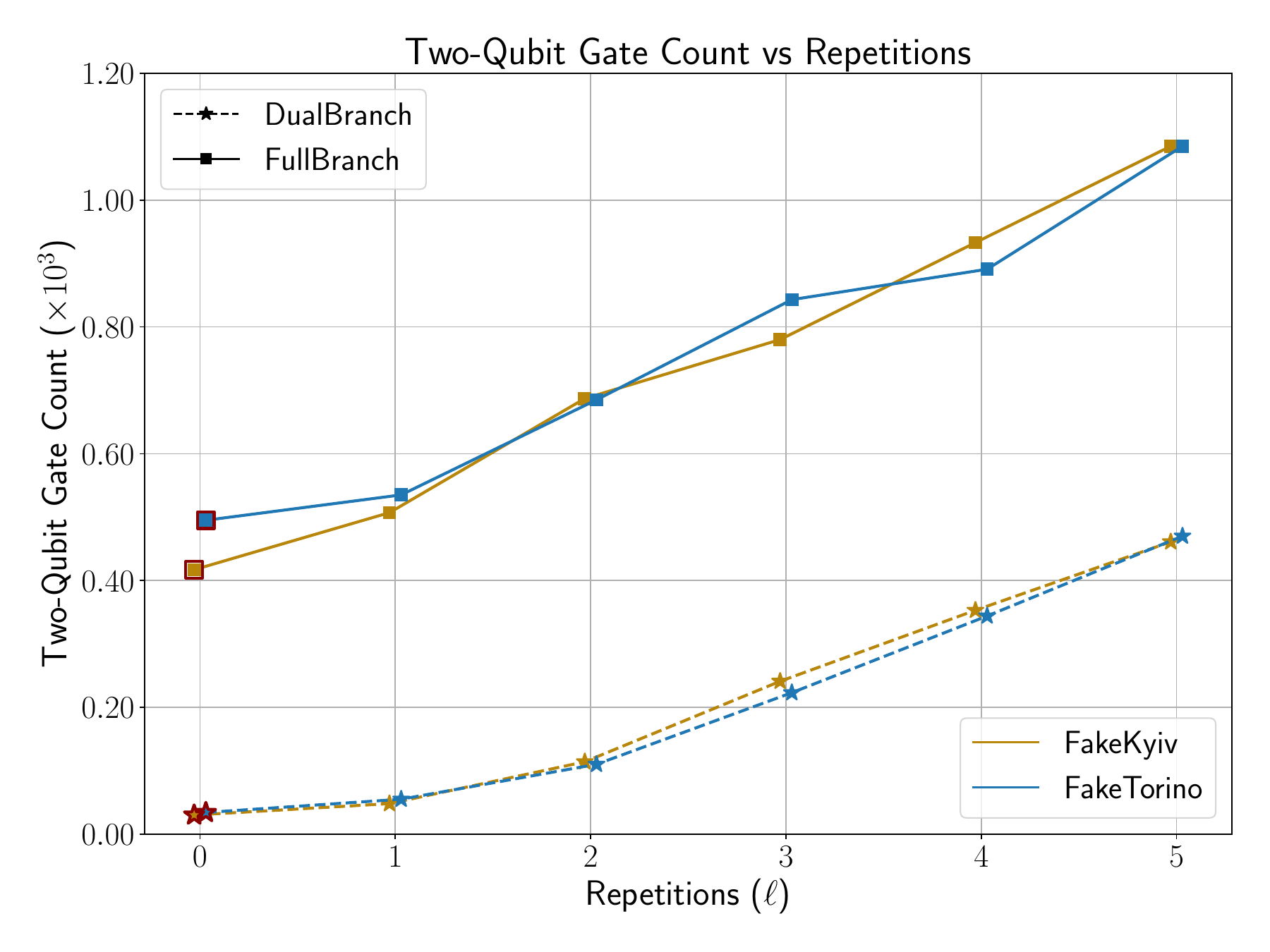}
        \caption{Total number of two-qubit gates (CZ and ECR)}
        \label{fig:ccount}
    \end{subfigure}
    \hfill
    \begin{subfigure}[t]{0.497\textwidth}
        \centering
        \includegraphics[width=\linewidth]{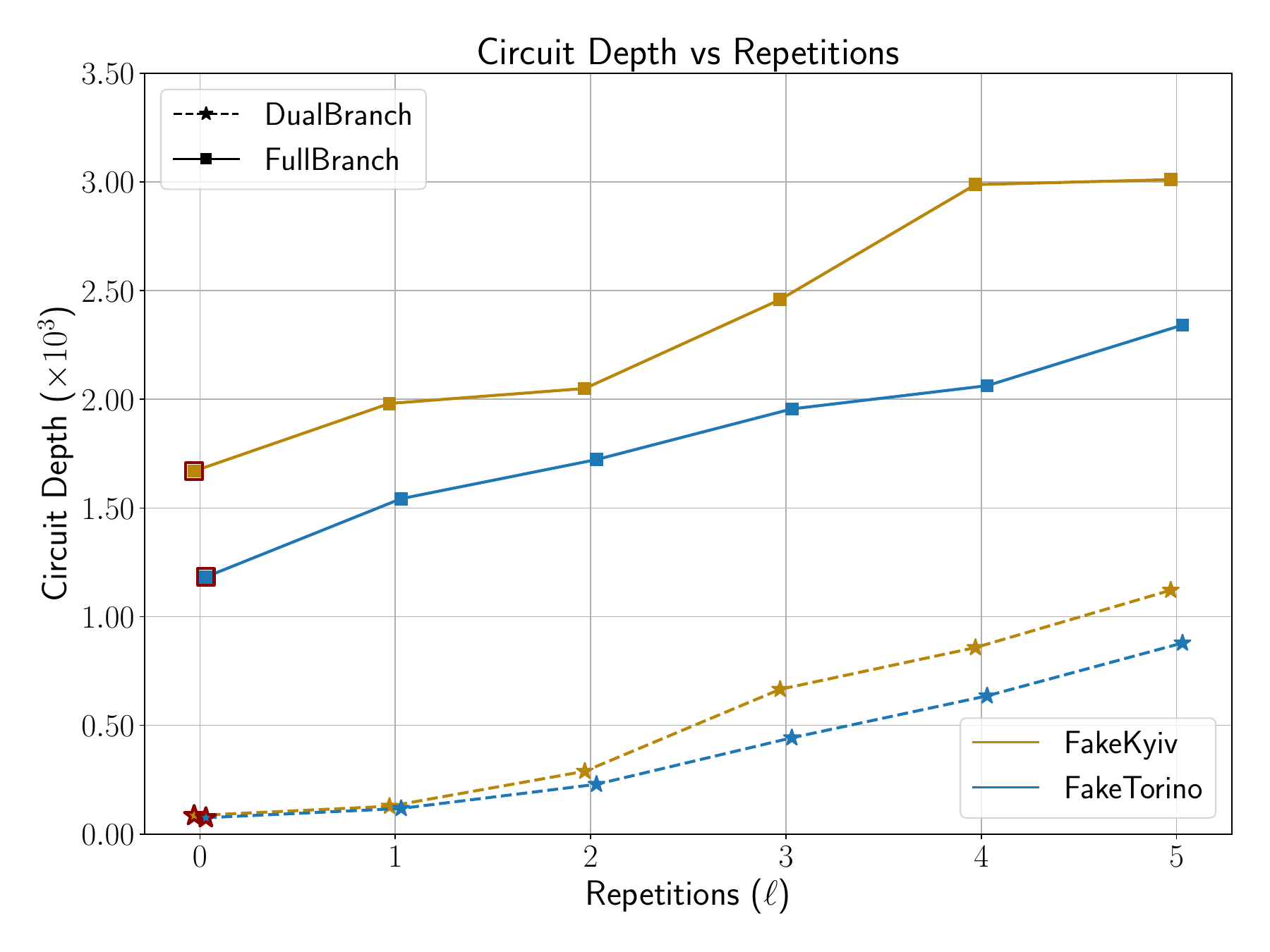}
        \caption{Circuit depth}
        \label{fig:cdepth}
    \end{subfigure}
    \caption{
    Evaluation of the two-qubit gates and circuit depth for different values of $\ell$, for Full and Dual-Branch models, and for different hardware. The number of repetitions selected in this work is $\ell=1$ for the Full model and $\ell=2$ for the Dual-Branch model.  
    }
    \label{fig:transpilation_count}
    \vspace{-10pt}
\end{figure}

To better understand the source of noise sensitivity, we extracted gate statistics from the transpiled circuits (Fig.~\ref{fig:transpilation_count}).

The analysis confirms that Dual-Branch circuits systematically require fewer two-qubit gates than Full circuits of comparable $\ell$ (Fig.~\ref{fig:ccount}). Since two-qubit gates are the dominant source of error in superconducting architectures, this reduction directly translates into higher state fidelities. Also, the gate depth is less than one third of the one required by the Full model for the values of $\ell$ considered in this work (Fig.~\ref{fig:cdepth}).

It is important to stress that the reported fidelities are conservative estimates:
\begin{itemize}
    \item No optimised transpilation strategies beyond the default Qiskit pass manager (with \texttt{$optimization\_level=4$}) were employed \citep{noise_compilation, noise_compilation2}.
    \item No circuit-level error mitigation (e.g., Pauli twirling \citep{Pauli_Twir} Clifford twirling  \citep{Clifford_Twir}) was applied.
    \item No post-processing techniques such as readout error mitigation or HAMMER \citep{HAMMER} were considered.
\end{itemize}

In realistic hardware runs, these methods are routinely applied and are known to significantly improve fidelities. Moreover, when training is performed directly on noisy hardware, model parameters may adapt to partially compensate for systematic error patterns.

\section{\textcolor{black}{Varying Number of Qubits}}
\label{app:h}

\begin{figure}[htbp]
    \centering
    \includegraphics[
        width=0.7\textwidth,
        trim=0 0 0 0mm,
        clip
    ]{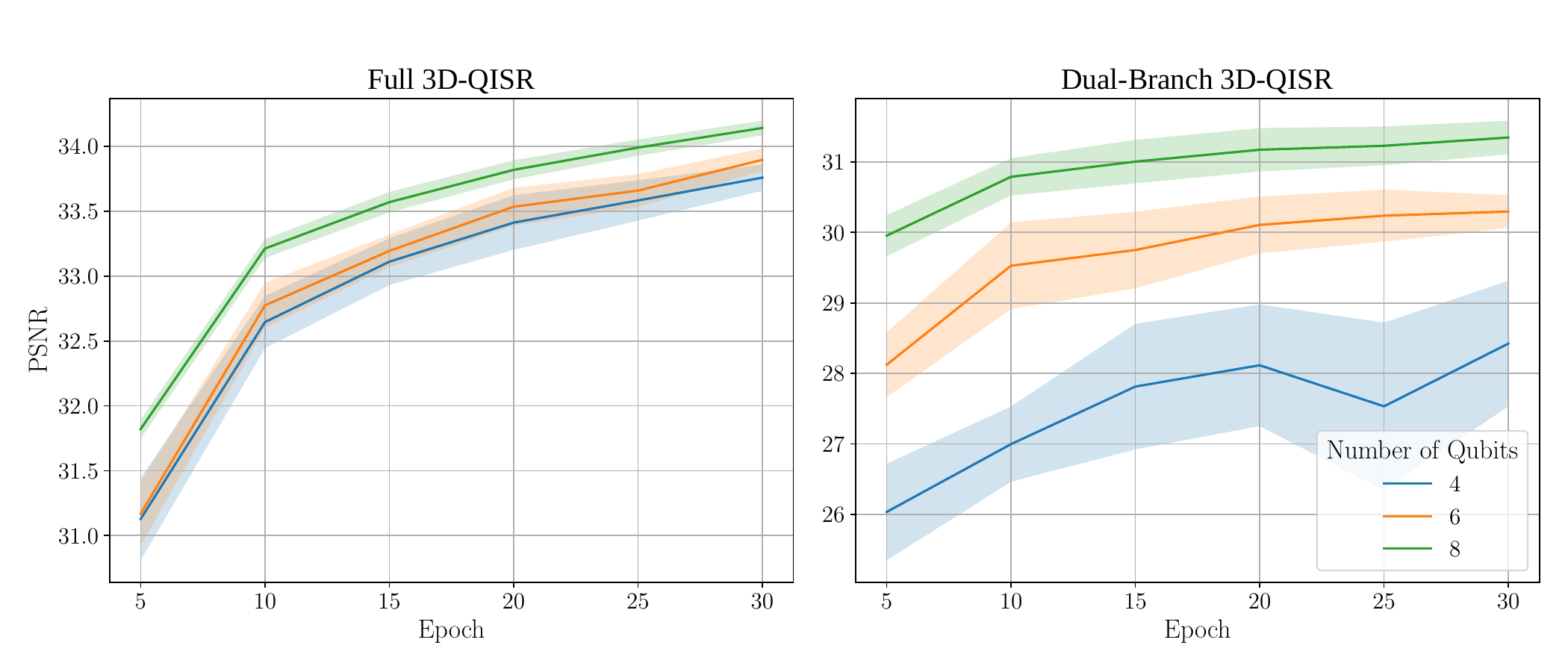}
    \caption{\textcolor{black}{
        Training curves (test PSNR over epochs) on the \textit{Lego} scene
        for different numbers of qubits. The left panel shows the
        \textbf{Full 3D-QISR} model, while the right panel shows the
        \textbf{Dual-Branch 3D-QISR} model. The shaded areas represent
        the mean $\pm$ standard deviation.
    }}
    \label{fig:app:ablation_qubits}
\end{figure}

\begin{figure*}[htbp]
    \centering

    \begin{minipage}[t]{0.32\textwidth}
        \centering
        \includegraphics[width=\textwidth]{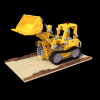}
        \subcaption*{(a) Full 3D-QISR, 4 qubits}
    \end{minipage}
    \hfill
    \begin{minipage}[t]{0.32\textwidth}
        \centering
        \includegraphics[width=\textwidth]{img_65.png}
        \subcaption*{(b) Full 3D-QISR, 6 qubits}
    \end{minipage}
    \hfill
    \begin{minipage}[t]{0.32\textwidth}
        \centering
        \includegraphics[width=\textwidth]{img_65.png}
        \subcaption*{(c) Full 3D-QISR, 8 qubits}
    \end{minipage}

    \vspace{0.8em}

    \begin{minipage}[t]{0.32\textwidth}
        \centering
        \includegraphics[width=\textwidth]{img_65.png}
        \subcaption*{(d) Dual-Branch 3D-QISR, 4 qubits}
    \end{minipage}
    \hfill
    \begin{minipage}[t]{0.32\textwidth}
        \centering
        \includegraphics[width=\textwidth]{img_65.png}
        \subcaption*{(e) Dual-Branch 3D-QISR, 6 qubits}
    \end{minipage}
    \hfill
    \begin{minipage}[t]{0.32\textwidth}
        \centering
        \includegraphics[width=\textwidth]{img_65.png}
        \subcaption*{(f) Dual-Branch 3D-QISR, 8 qubits}
    \end{minipage}

    \caption{\textcolor{black}{Rendered output for the ablation study across models and qubit counts. The first row shows Full 3D-QISR results for 4, 6, and 8 qubits. The second row shows the corresponding Dual-Branch 3D-QISR reconstructions.}}
    \label{fig:app:ablation-grid}
\end{figure*}

We evaluate how the number of qubits $n$ impacts the performance of our method on the \textit{Lego} scene (see Fig.~\ref{fig:app:ablation_qubits}). 
While the Full 3D-QISR model shows competitive performance for $n=4$ and $n=6$, the Dual-Branch model requires $8$ qubits to perform comparably to the classical counterpart. 
Notice that DB 3D-QISR with $n=6$ has the same number of amplitudes encoded as the Full model with $n=4$. 
The latter has more parameters but lower performance, probably due to the lower expressivity of the model (e.g.,~due to the different entangling layer structure). 
On the other hand, it is interesting to note that the classical parameter counts in DB 3D-QISR for $n=4, 6, 8$ are $290k$, $293k$, and $297k$,  respectively. 
This suggests that the dramatic increase in performance depends predominantly on the increase in the representation capacity of the quantum component. 
Finally, we visualise views for the tested models in Fig.~\ref{fig:app:ablation-grid}. 
Novel-view renderings by Full 3D-QISR show close to no perceptual difference in the reconstruction quality for different $n$.

\section{\textcolor{black}{Estimated execution time on \textit{IBM\_Torino}}}
\label{app:ext-time}
Starting from the analysis in App.~\ref{app:noise}, we provide the expected execution time on the \textit{IBM\_Torino} gate-based quantum device \citep{qiskit_fake_provider}. In practice, the transpilation process for a specific hardware produces the so-called Instruction Set Architecture (ISA) circuit representation, where each gate has a specific time duration. In this way, we are able to estimate the duration of a circuit execution on a target hardware.  

In practice, the execution time provides a lower bound on the time that would be required to execute our proposed models on this specific quantum hardware. In real-world applications, the total time would increase, as in hybrid algorithms, quantum and classical devices must exchange information. Also, the time on a given device depends on additional factors, such as compilation and transpilation pipelines. 
Moreover, note that different quantum hardware can have substantially different execution times.

\begin{figure}[htbp]
    \centering
    \begin{minipage}[t]{0.49\textwidth}
        \centering
        \includegraphics[width=\textwidth, trim=0 0 0 10pt, clip]{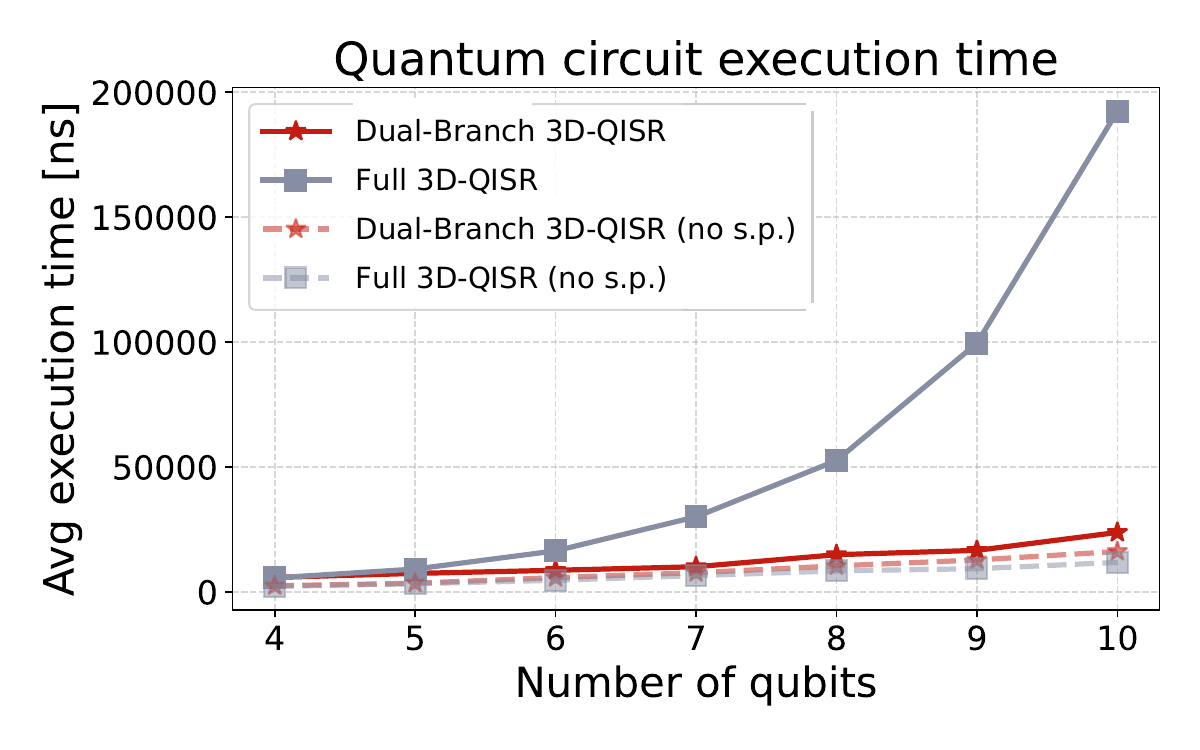}
        \subcaption*{(a) Average execution time for $100$ circuits with qubit counts from $4$ to $10$. Dashed lines indicate post–state-preparation time.}
    \end{minipage}
    \hfill
    \begin{minipage}[t]{0.49\textwidth}
        \centering
        \includegraphics[width=\textwidth, trim=0 0 0 10pt, clip]{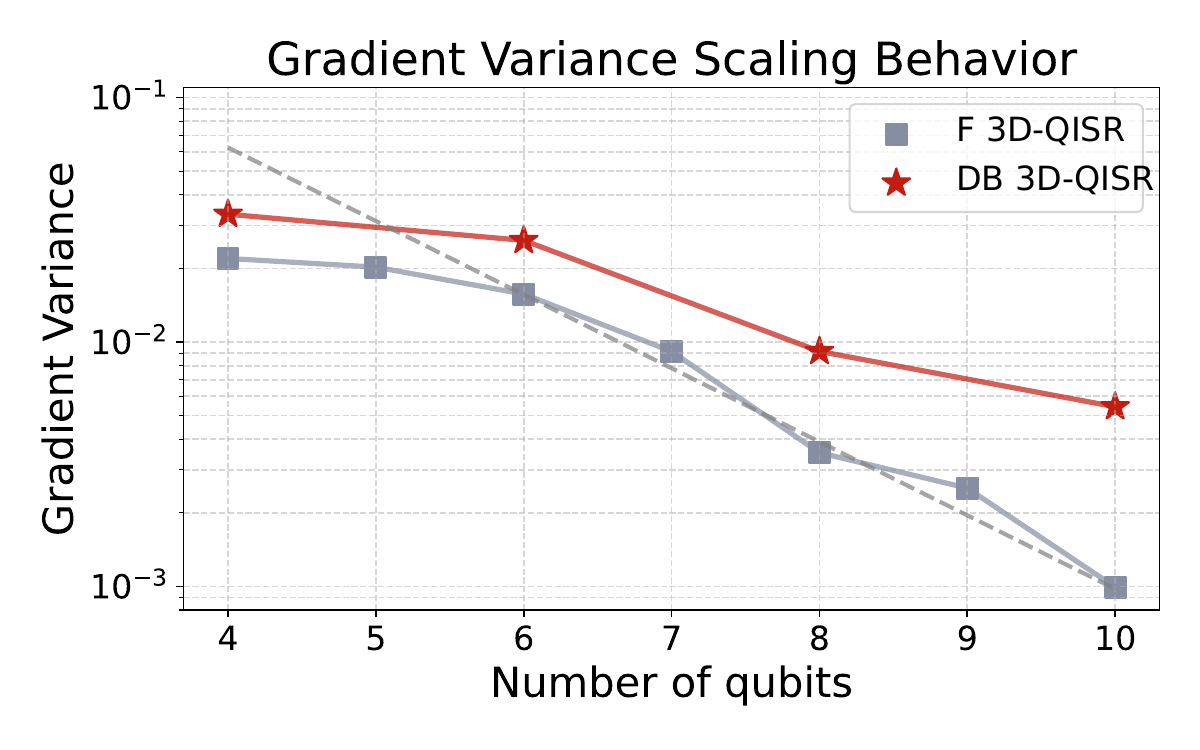}
        \subcaption*{(b) Average gradient variance of the proposed models for Blender scenes. The dashed line references the value $2^{-n}$.} 
    \end{minipage}%
    \caption{\textcolor{black}{Comparison of gradient variance and quantum execution time for the evaluated models. Panel (a) presents expected execution times on \textit{IBM\_Torino} for circuits of increasing size, while panel (b) reports the variance of gradients on simulated hardware across scenes from the Blender dataset.}}
    \label{fig:app:gradvar-exttime}
\end{figure}

Fig.~\ref{fig:app:gradvar-exttime} (a) shows the average time required to execute the considered quantum circuits with $\ell=1$ for the Full 3D-QISR, and $\ell=2$ for the Dual-Branch model. The number of qubits $n$ varies from $4$ to $10$ to assess scalability. 
Note that for the Full 3D-QISR model, most of the execution time depends on the state preparation. In particular, by using exact state preparation (i.e.~the amplitude embedding routine from \textit{qiskit}, we observe an exponential increase in the total time. 
This is expected, as amplitude embedding requires, in general, up to $2^n$ base operations. 
On the other hand, DB 3D-QISR requires much less time, as the partial amplitude embedding requires exponentially fewer operations than the standard amplitude embedding. 

In practice, in our use case for $n=8$, the DB model requires $3.7\times$ less time than Full ($1.48\times 10^4$ns vs  $5.25\times 10^4$ns) for one execution. However, note that the execution time depends on the transpiled circuit to be executed on the hardware: different hardware specifics can drastically reduce the number of physical gates, and therefore the total time (e.g. assuming $8$ qubits with full connectivity on \textit{IBM$\_$Torino} could reduce the total execution time by an order of magnitude).
\section{\textcolor{black}{Gradient Estimation}}
\label{app:j}
We now evaluate empirically the initial gradient variance 
of our proposed models. A well-known limitation on QML model scaling is the so-called Barren Plateau (BP) \citep{BP}, i.e.~an effect during QNN training when the variance of the gradient decreases exponentially with the size of the system 
BPs are believed to increase the number of shots required to estimate the gradient and, therefore, also the total cost of the algorithm with the number of qubits of the system. 
BPs are correlated with the 
expressivity 
of the ansatz: In general, more expressive models are more prone to exhibit lower gradient variance. 
In Fig.~\ref{fig:app:gradvar-exttime}-(b), we evaluate 
the average gradient variance over $100$ random initialisations. 
As BP depends on the task under consideration (and the associated energy optimisation landscape), for each initialisation, we select $100$ batches of size $64$ from a task in the Blender dataset and evaluate the average gradient variance. 
By doing so, we estimate the gradient in the same setting as the one for the main experiments. 
We fix the parameter $\ell=1$ for Full and $\ell=2$ for Dual-Branch, and select the number of qubits from the set $n\in\{4, 5, 6, 7, 8, 9, 10\}$.
The Full 3D-QISR architecture exhibits behaviour implying BPs, i.e.,~an exponential decay of the variance. 
This is consistent with our observations on the limited scalability of this model. 
However, BPs can become a limitation for higher $n$: Even if the gradient estimation requires an exponential number of shots, in our experimental setting $2^n=256$, which is still feasible. 
Many quantum hardware producers use pipelines optimised for a higher number of shots (e.g.~the default shot number of \textit{IBM\_Torino} is $5000$, which is more than enough to provide a faithful estimation of the gradient with $8$ qubits). 
On the other hand, the DB architecture, due to its lower
expressivity, exhibits better scaling properties. 
For $n=8$, the gradient variance is approximately three times higher than for the Full counterpart ($9.28\times 10^{-3}$ vs~$3.41\times 10^{-3}$), and approximately four times for $n=10$ ($4.13\times 10^{-4}$ vs $1.12\times 10^{-4}$). 
Note that in our experiments on the DB model, we always assume that the number of qubits to encode positional and view-dependent features is the same. For this reason, we only consider even values of $n$.
\section{\textcolor{black}{Mesh Extraction}}
\label{app:l}
\begin{figure}[htbp]
    \centering

    \begin{minipage}[t]{0.245\textwidth}
        \centering
        \includegraphics[width=\textwidth,
                         trim=400px 0 400px 0, clip]{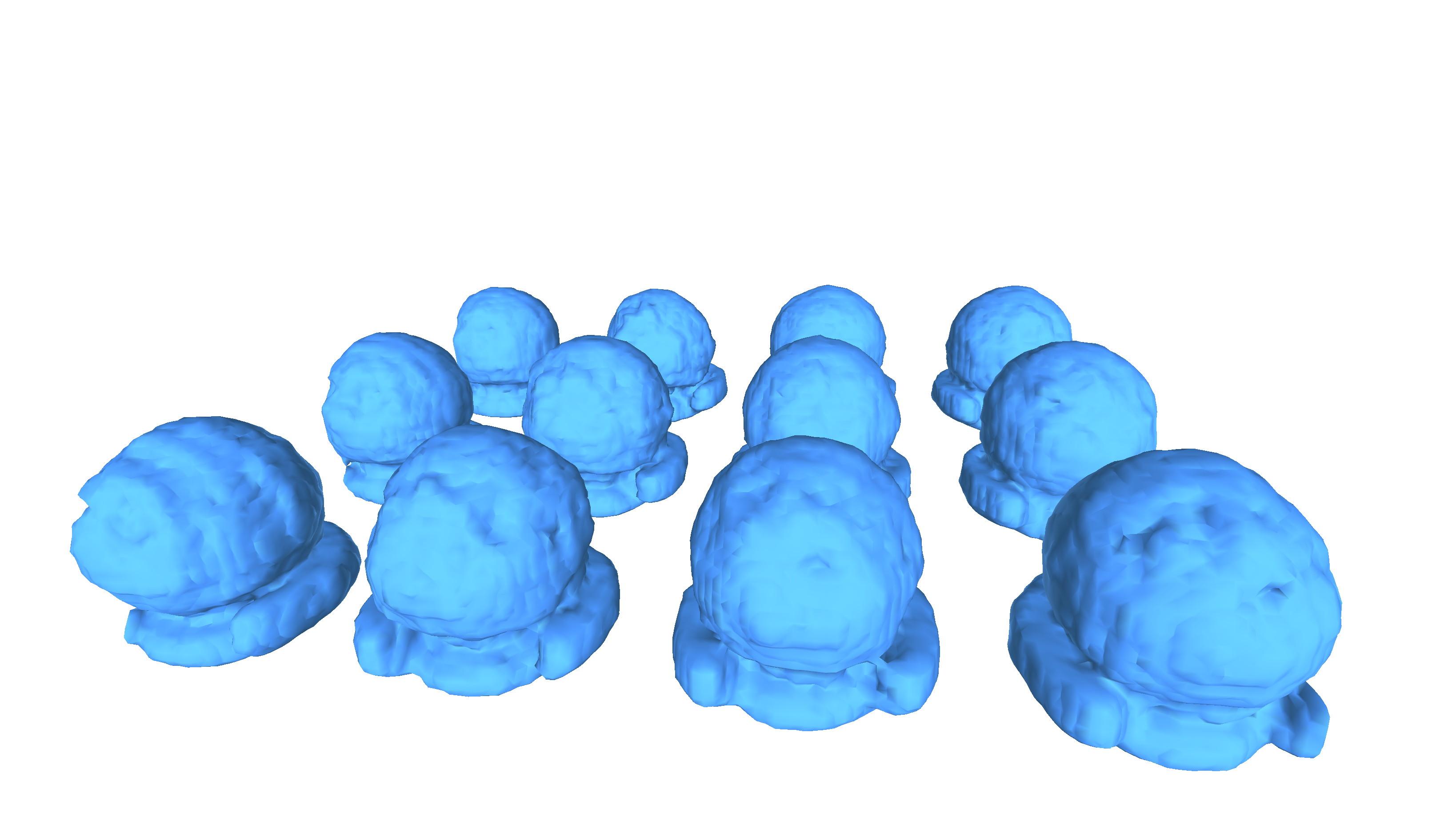}
        \subcaption*{Materials}
    \end{minipage}
    \hfill
    \begin{minipage}[t]{0.245\textwidth}
        \centering
        \includegraphics[width=\textwidth,
                         trim=400px 0 400px 0, clip]{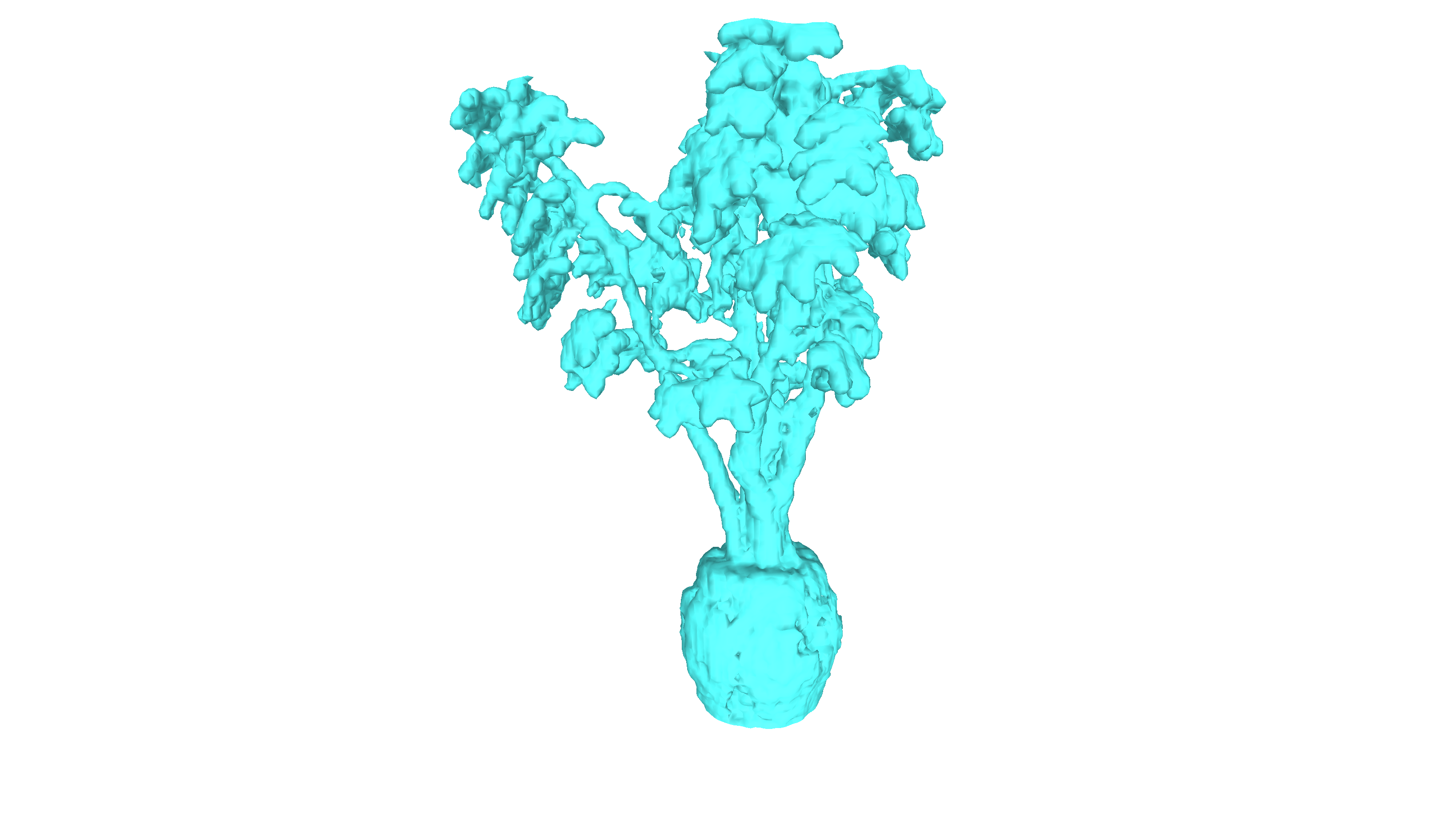}
        \subcaption*{Ficus}
    \end{minipage}
    \hfill
    \begin{minipage}[t]{0.245\textwidth}
        \centering
        \includegraphics[width=\textwidth,
                         trim=400px 0 400px 0, clip]{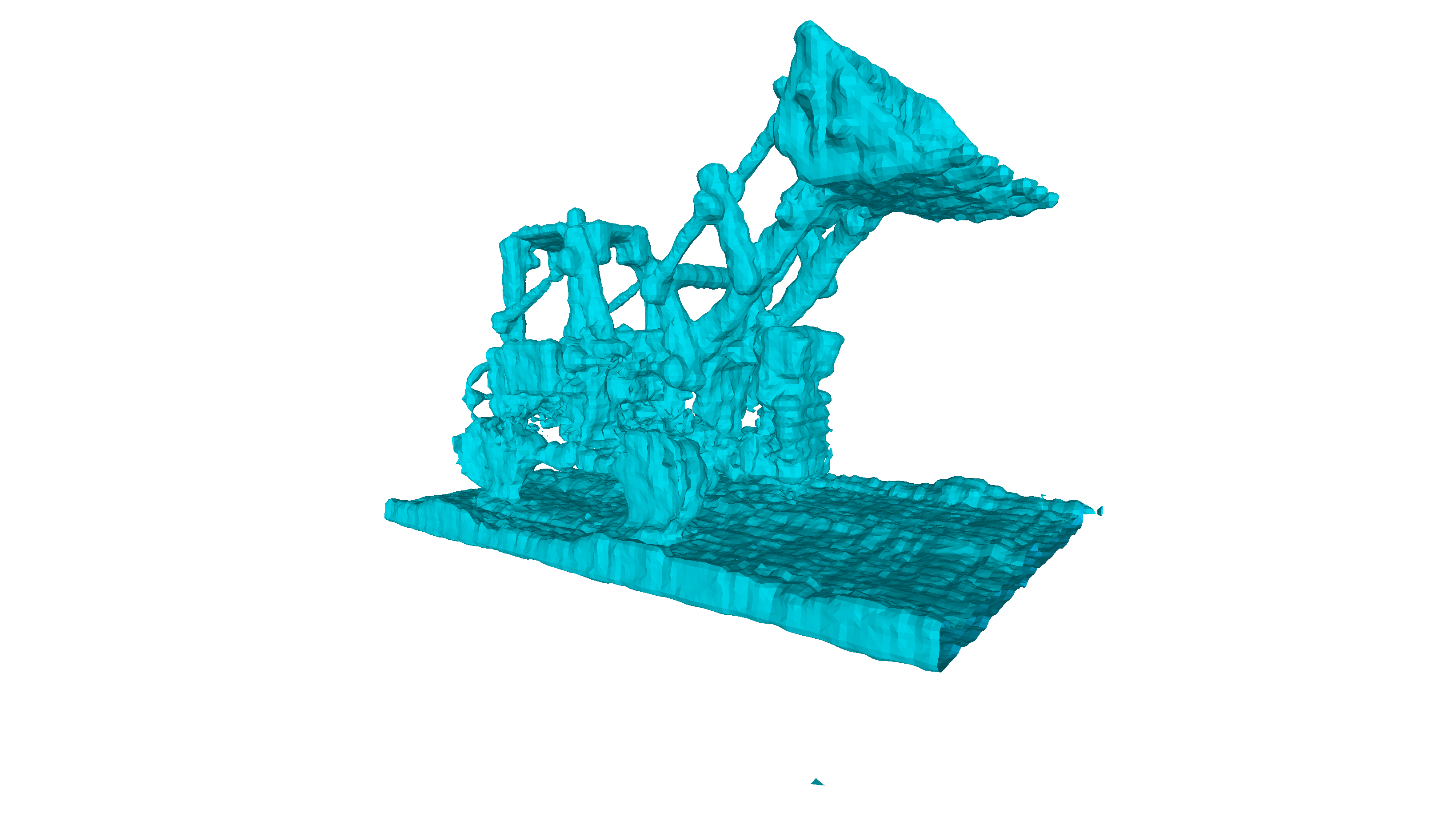}
        \subcaption*{Lego}
    \end{minipage}
    \hfill
    \begin{minipage}[t]{0.245\textwidth}
        \centering
        \includegraphics[width=\textwidth,
                         trim=400px 0 400px 0, clip]{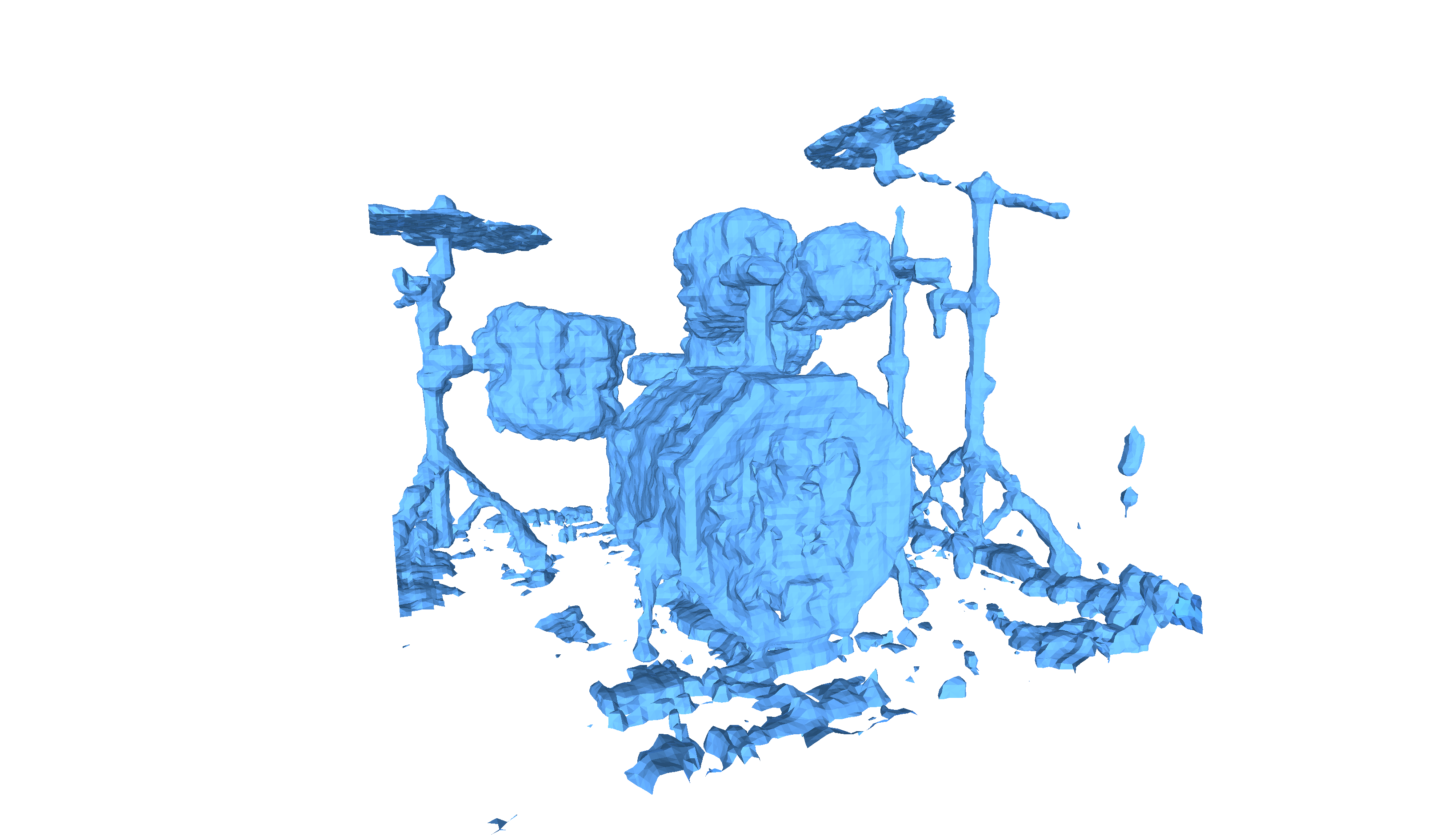}
        \subcaption*{Drums}
    \end{minipage}

    \caption{\textcolor{black}{MeshLab rendering of the 3D meshes extracted using Marching Cubes from the trained Full 3D-QISR model.}}
    \label{fig:app:mesh} 
\end{figure}

Similar to classical NeRF, 3D-QISR can be used to extract meshes from the trained models. 
For example, we provide renderings of extracted 3D meshes in Fig.~\ref{fig:app:mesh}. 
Each image is a render of the 3D object extracted from a trained Full 3D-QISR model with $8$ qubits using the Marching Cubes (MC) algorithm \citep{MCubes}. 
The MC algorithm is implemented in the Python library \textit{skimage}. 
The visualised objects are obtained with a resolution of $128^3$, and a sigma threshold of $10$. 
Finally, the final rendering is obtained using the MeshLab software \citep{meshlab}. 

\section{\textcolor{black}{Enhancing 3D-QISR with Conical Frustum and IPE}}
\label{app:m}

NeRF established a new benchmark for novel-view synthesis. After its introduction, a wide range of extensions subsequently emerged, each targeting improvements in rendering efficiency, generalisation, or robustness under challenging imaging conditions—for example, mip-NeRF \citep{mip-NERF} for multi-scale rendering.

Although these more recent approaches vary in architectural detail, many retain the same foundational structure: one or more multilayer perceptrons are trained to approximate a volumetric radiance field from posed 2D observations. Subsequent works primarily differ in how they parametrise and sample the underlying volume. Among these, mip-NeRF \citep{mip-NERF} replaces ray-based point sampling with a multi-scale, cone-based formulation that models the integration of signals across spatial frequencies when compared to the original NeRF, yielding improved anti-aliasing and higher fidelity under varying resolutions.

This paper shows that augmenting NeRF with quantum neural networks can improve novel-view synthesis quality while simultaneously reducing the model size. 
Because the quantum components operate at the level of the underlying continuous signal representation (i.e.,~the underlying MLP), the resulting hybrid architecture remains compatible with a wide class of NeRF variants. 
This further motivates an investigation of whether the same quantum enhancements can be extended to more advanced formulations.

To this end, we examine the integration of QNN modules into mip-NeRF \citep{mip-NERF}. Compared to the original NeRF design, mip-NeRF introduces two central modifications: (i) it replaces infinitesimal point samples with conical frustums, assigning each sample a finite footprint along the viewing ray; and (ii) it integrates the radiance field over these regions using a multiscale representation derived from integrated positional encoding (IPE). 

We evaluate both classical mip-NeRF and a ``Quantum-Mip-NeRF'' (QMip-NeRF) on the same setting used for the main experiment (single-scale Blender dataset downscaled to $100\times 100$ pixels). The code for conical frustum sampling and IPE was adapted from \citep{multinerf}. No other component (model structure, etc.) was modified. 
We train each model for up to 15 epochs (requiring approximately $15$ hours per scene), and we repeated each experiment $5$ times. We report the average PSNR for the scenes of the Blender dataset in Tab.~\ref{tab:mipvalues}. 

\begin{figure*}[htbp]
    \centering

    \begin{minipage}[t]{0.95\textwidth}
        \centering
        \includegraphics[width=\textwidth]{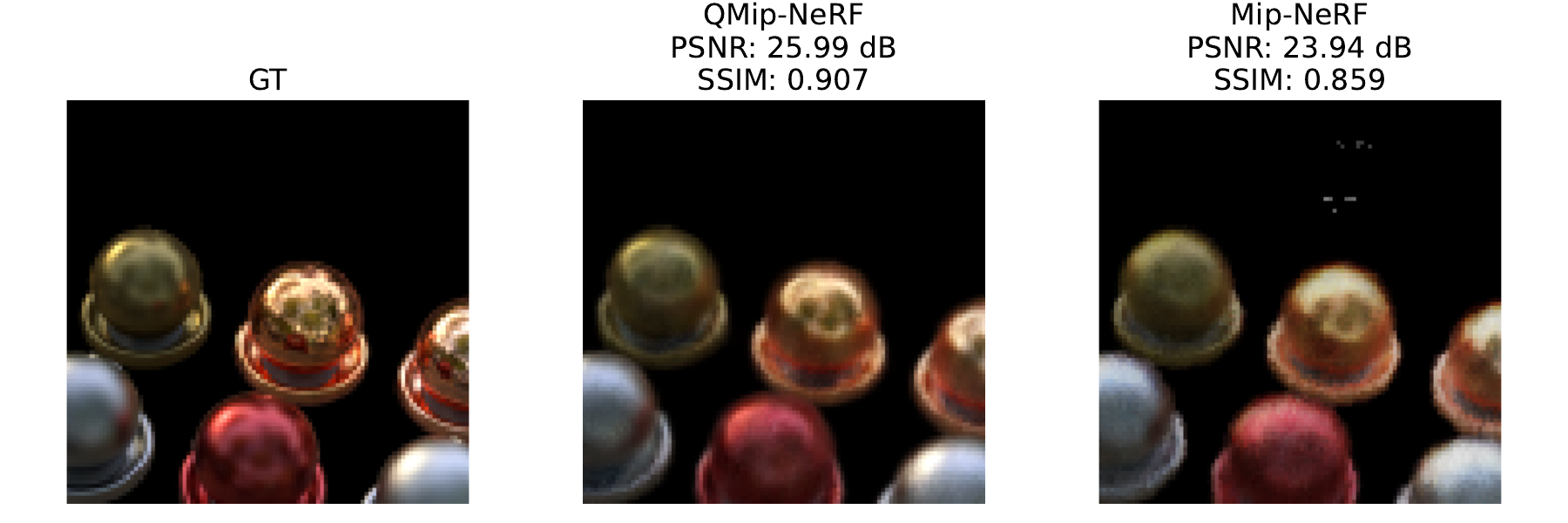}
    \end{minipage}
    \hfill
    \begin{minipage}[t]{0.95\textwidth}
        \centering
        \includegraphics[width=\textwidth]{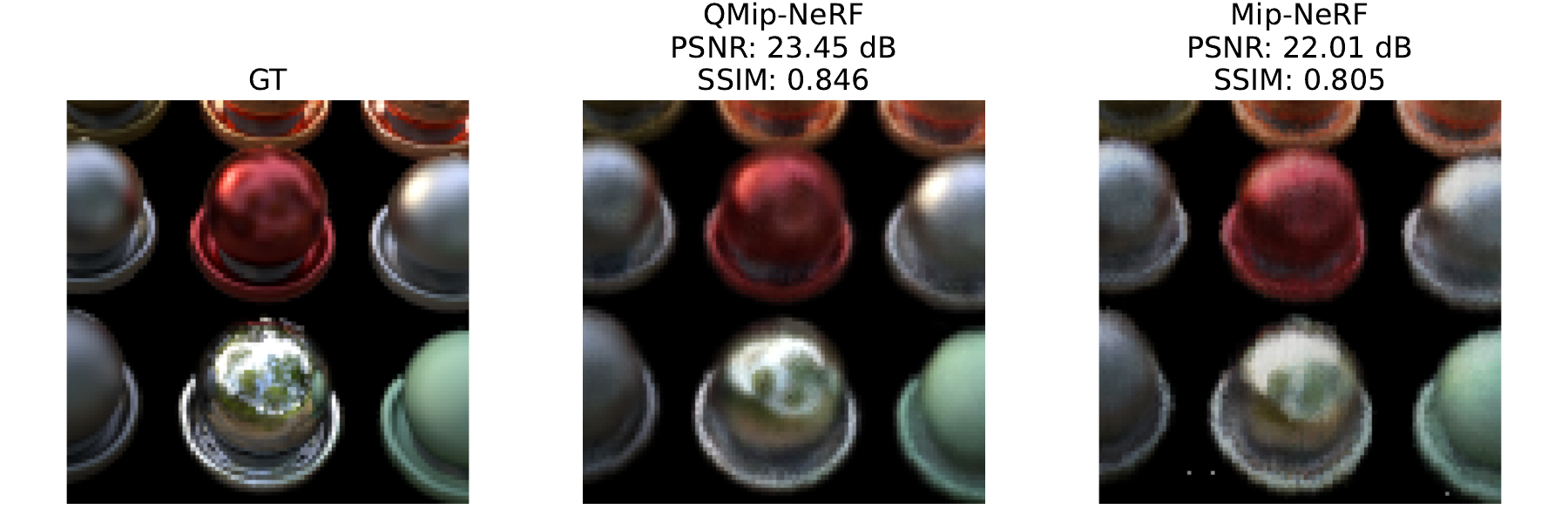}
    \end{minipage}
    \hfill
    \begin{minipage}[t]{0.95\textwidth}
        \centering
        \includegraphics[width=\textwidth]{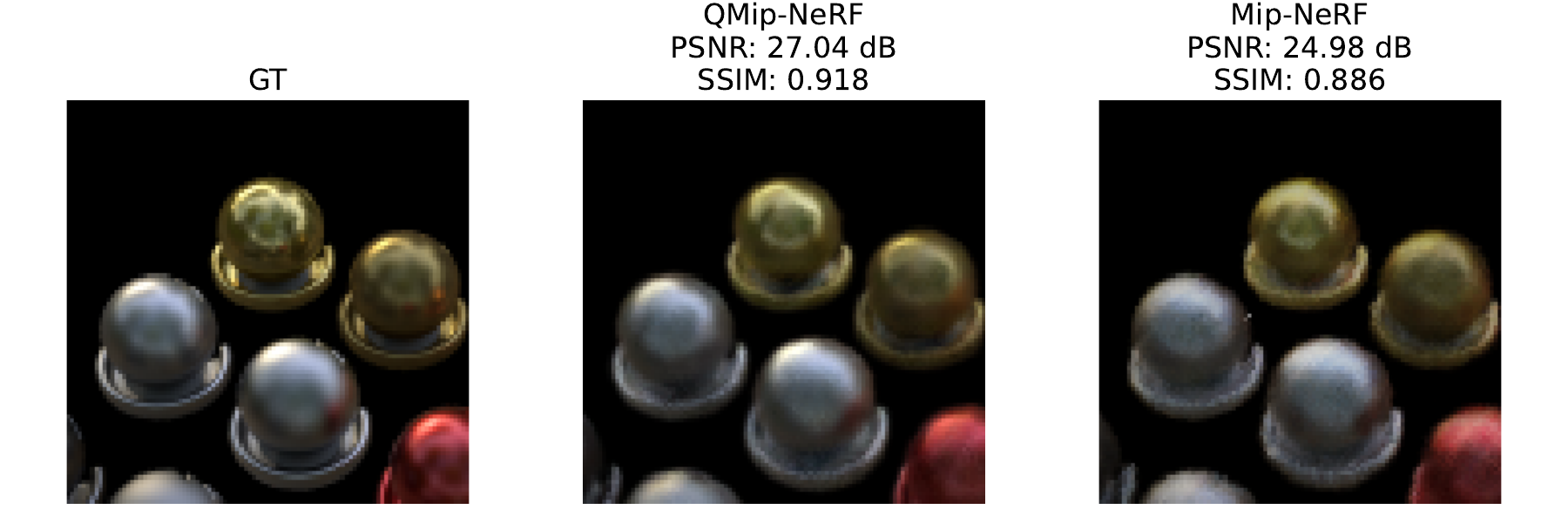}
    \end{minipage}
\caption{\color{black}{Visualisation of mip-NeRF variants obtained by inference at a higher resolution ($200\times 200$ pixels) on the \textit{materials} scene.}} 
    \label{fig:mip-var}
\end{figure*}

Note that---despite not being trained on the multi-scale Blender dataset---the considered models can be used to represent scenes with higher resolution (i.e., by using a different scale). In Fig.~\ref{fig:mip-var}, we provide visualisations on the \textit{materials} scene to highlight differences in high-frequency details. 

\begin{table*}[h!]
\caption{\color{black}{PSNR after 15 epochs for ``QMip-NeRF'' and classical mip-NeRF.}}
\label{tab:mipvalues} 
\centering
\begin{tabular}{lccccc}
\toprule
 & \textit{Materials} & \textit{Ficus} & \textit{Lego} & \textit{Drums} & Average \\
\midrule
QMip-NeRF & 
$32.43\pm0.25$ &
$29.59\pm0.12$ &
$33.25\pm0.11$ &
$27.58\pm0.11$ &
$30.71\pm 0.15$ \\
mip-NeRF &
$30.53\pm0.69$ &
$29.81\pm0.34$ &
$32.73\pm0.27$ &
$27.09\pm0.16$ &
$30.04\pm0.37$ \\
\bottomrule
\end{tabular}

\end{table*}

\end{document}